\RequirePackage{fix-cm}
\documentclass[twocolumn]{svjour3}          
\smartqed  
\usepackage{graphicx}
\usepackage{amsfonts}
\usepackage{amsmath}
\usepackage{enumitem}
\usepackage{array}
\usepackage[table]{xcolor}
\usepackage{url}
\usepackage{natbib}
\usepackage{mathptmx}      
\usepackage{booktabs}
\usepackage{multirow}

\newcommand{\sect}[1]{Sect.~\ref{sec:#1}}
\newcommand{\fig}[1]{Fig.~\ref{fig:#1}}
\newcommand{\tab}[1]{Table~\ref{tab:#1}}
\newcommand{\eqn}[1]{(\ref{eq:#1})}

\newcommand{\given}{\,|\,}
\newcommand{\Given}{\,\Big\vert\,}
\newcommand{\kldiv}[2]{\mathit{KL}\left[ #1 \,\Big\rvert\Big\rvert\, #2 \right]}
\newcommand{\iougt}{$\mathrm{iou} \given \theta$}
\newcommand{\itiougt}{$\mathit{iou} \given \theta$}

\renewcommand{\paragraph}[1]{\vspace{4pt}\noindent\textbf{#1}~~}

\begin{document}

\title{
Learning single-image 3D reconstruction by generative modelling of shape, pose and shading
}


\author{Paul Henderson         \and
        Vittorio Ferrari
}


\institute{P. Henderson \at
	      Institute of Science and Technology (IST) Austria \\
              \email{paul@pmh47.net}           
           \and
           V. Ferrari \at
              Google Research, Z\"{u}rich, Switzerland \\
	      \email{vittoferrari@google.com}
}

\date{~}

\maketitle

\begin{abstract}
\begin{sloppypar}
We present a unified framework tackling two problems: class-specific 3D reconstruction from a single image, and generation of new 3D shape samples.
These tasks have received considerable attention recently;
however, most existing approaches rely on 3D supervision, annotation of 2D images with keypoints or poses, and/or training with multiple views of each object instance.
Our framework is very general:  it can be trained in similar settings to existing approaches, while also supporting weaker supervision. Importantly, it can be trained purely from 2D images, without pose annotations, and with only a single view per instance.
We employ meshes as an output representation, instead of voxels used in most prior work.
This allows us to reason over lighting parameters and exploit shading information during training, which previous 2D-supervised methods cannot.
Thus, our method can learn to generate and reconstruct concave object classes.
We evaluate our approach in various settings, showing that:
(i) it learns to disentangle shape from pose and lighting;
(ii) using shading in the loss improves performance compared to just silhouettes;
(iii) when using a standard single white light, our model outperforms state-of-the-art 2D-supervised methods, both with and without pose supervision, thanks to exploiting shading cues;
(iv) performance improves further when using multiple coloured lights, even approaching that of state-of-the-art 3D-supervised methods;
(v) shapes produced by our model capture smooth surfaces and fine details better than voxel-based approaches; and
(vi) our approach supports concave classes such as bathtubs and sofas, which methods based on silhouettes cannot learn.
\end{sloppypar}

\end{abstract}

\section{Introduction}
\label{sec:intro}

\begin{figure*}
  \centering
  \includegraphics[width=0.85\linewidth]{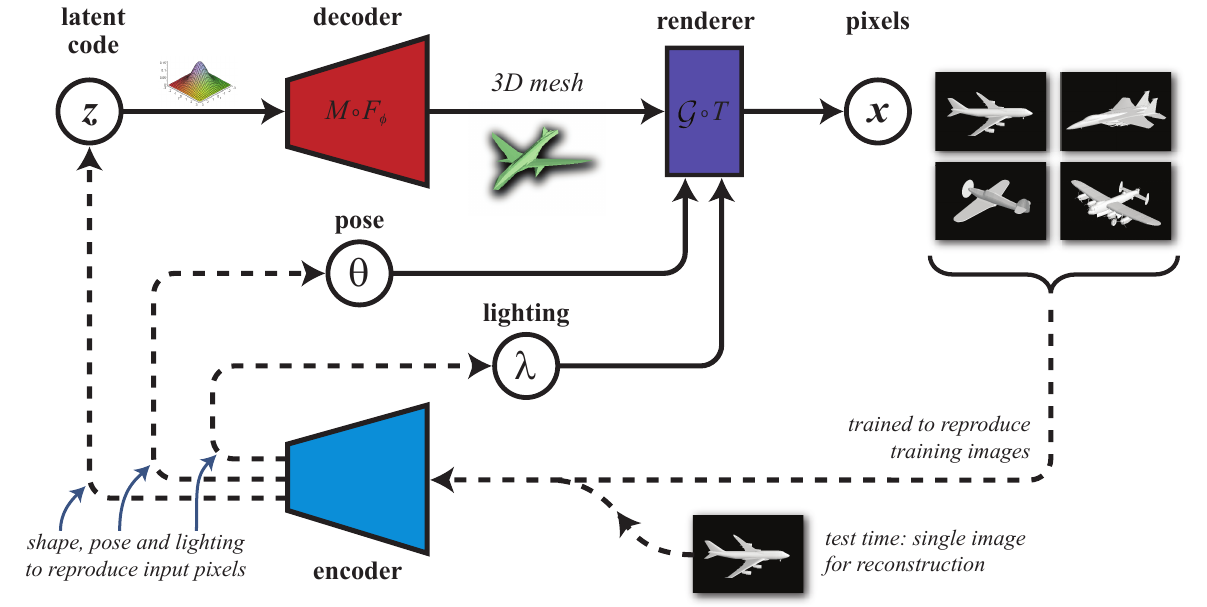}
  \caption{
Given only unannotated 2D images as training data, our model learns (1) to reconstruct and predict the pose of 3D meshes from a single test image, and (2) to generate new 3D mesh samples.
 The generative process (solid arrows) samples a Gaussian embedding, decodes this to a 3D mesh, renders the resulting mesh, and finally adds Gaussian noise.
It is trained end-to-end to reconstruct input images (dashed arrows), via an encoder network that learns to predict and disentangle shape, pose, and lighting.
  The renderer produces lit, shaded RGB images, allowing us to exploit shading cues in the reconstruction loss.
}
  \label{fig:flow}
\end{figure*}

Reconstructing 3D objects from 2D images is a long-standing research area in computer vision.
While traditional methods rely on multiple images of the same object instance~\citep{seitz2006comparison,furukawa15cgft,broadhurst01iccv,laurentini94pami,debonet99iccv,gargallo07accv,liu10cvpr}, there has recently been a surge of interest in learning-based methods that can infer 3D structure from a single image, assuming that it shows an object of a class seen during training (e.g. \citealp{fan17cvpr,choy16eccv,yan16nips}; see \sect{relwork/recon}).
A related problem to reconstruction is that of generating new 3D shapes from a given object class \textit{a priori}, i.e. without conditioning on an image.
Again, there have recently been several works that apply deep learning techniques to this task (e.g. \citealp{wu16nips,zou17iccv,gadelha173dv}; see \sect{relwork/gen}).

Learning-based methods for single-image reconstruction are motivated by the fact that the task is inherently ambiguous: many different shapes project to give the same pixels, for example due to self-occlusion.
Hence, we must rely on prior knowledge capturing what shapes are likely to occur.
However, most reconstruction methods are trained discriminatively to predict complete shapes from images---they do not represent their prior knowledge about object shapes as an explicit distribution that can generate shapes \textit{a priori}.
In this work, we take a generative approach to reconstruction, where we learn an explicit  prior model of 3D shapes, and integrate this with a renderer to model the image formation process.
Inference over this joint model allows us to find the most likely 3D shape for a given image.

Most learning-based methods for reconstruction and generation rely on strong supervision.
For generation \citep[e.g.][]{wu16nips,zou17iccv},
this means learning from large collections of manually constructed 3D shapes, typically ShapeNet~\citep{shapenet15arxiv} or ModelNet~\citep{wu15cvpr-shapenets}.
For reconstruction \citep[e.g.][]{choy16eccv,fan17cvpr,richter18cvpr},
it means learning from images paired with aligned 3D meshes, which is very expensive supervision to obtain~\citep{yang18eccv}.
%
While a few methods do not rely on 3D ground-truth, they still require keypoint annotations on the 2D training images~\citep{vicente14cvpr,kar15cvpr,kanazawa18eccv}, and/or multiple views for each object instance, often with pose annotations~\citep{yan16nips,wiles17bmvc,kato18cvpr,tulsiani18cvpr-mvc,insafutdinov18nips}.
In this paper, we consider the more challenging setting where we only have access to unannotated 2D images for training, without ground-truth pose, keypoints, or 3D shape, and with a single view per object instance.

\begin{sloppypar}
It is well known that \textit{shading} provides an important cue for 3D understanding~\citep{horn75pcv}.
It allows determination of surface orientations, if the lighting and material characteristics are known; this has been explored in numerous works on shape-from-shading over the years~\citep{horn75pcv,zhang99pami,barron15pami}.
Unlike learning-based approaches, these methods can only reconstruct non-occluded parts of an object, and achieving good results requires strong priors~\citep{barron15pami}.
Conversely, existing learning-based generation and reconstruction methods can reason over occluded or visually-ambiguous areas, but do not leverage shading information in their loss.
Furthermore, the majority use voxel grids or point clouds as an output representation.
Voxels are easy to work with, but cannot explicitly model non-axis-aligned surfaces, while point clouds do not represent surfaces explicitly at all.
In both cases, this limits the usefulness of shading cues.
To exploit shading information in a learning-based approach, we therefore need to move to a different representation; a natural choice is 3D \textit{meshes}.
Meshes are ubiquitous in computer graphics, and have desirable properties for our task: they can represent surfaces of arbitrary orientation and dimensions at fixed cost, and are able to capture fine details.
Thus, they avoid the visually displeasing `blocky' reconstructions that result from voxels.
We also go beyond monochromatic light, considering the case of coloured directional lighting; this provides even stronger shading cues when combined with arbitrarily-oriented mesh surfaces.
Moreover, our model explicitly reasons over the lighting parameters, jointly with the object shape, allowing it to exploit shading information even in cases where the lighting parameters are unknown---which classical shape-from-shading methods cannot.
\end{sloppypar}

\begin{figure}
  \includegraphics[width=\linewidth]{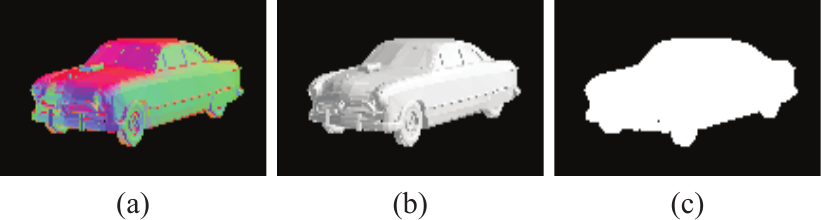}
  \caption{
  Lighting: Coloured directional lighting (a) provides strong cues for surface orientation; white light (b) provides less information; silhouettes (c) provide none at all. Our model is able to exploit the shading information from coloured or white lighting.
  }
  \label{fig:lighting}
\end{figure}

In this paper, we present a unified framework for both reconstruction and generation of 3D shapes, that is trained to model 3D meshes using only 2D supervision (\fig{flow}).
Our framework is very general, and can be trained in similar settings to existing models~\citep{tulsiani17cvpr,yan16nips,wiles17bmvc,tulsiani18cvpr-mvc}, while also supporting weaker supervision scenarios.
It allows:
\begin{itemize}
\item
use of different \textbf{mesh parameterisations}, which lets us incorporate useful modeling priors such as smoothness or composition from primitives
\item
    exploitation of \textbf{shading cues} due to monochromatic or coloured directional lighting (\fig{lighting}), letting us discover concave structures that silhouette-based methods cannot~\citep{gadelha173dv,tulsiani17cvpr,tulsiani18cvpr-mvc,yan16nips,soltani17cvpr}.
\item
training with \textbf{varying degrees of supervision}: single or multiple views per instance, with or without ground-truth pose annotations
\end{itemize}

\begin{sloppypar}
To achieve this, we design a probabilistic generative model that captures the full image formation process, whereby the shape of a 3D mesh, its pose, and incident lighting are first sampled independently, then a 2D rendering is produced from these (\sect{generative}).
We use stochastic gradient variational Bayes for training~\citep{kingma14iclr,rezende14icml} (\sect{training}).
This involves learning an \textit{inference network} that can predict 3D shape, pose and lighting from a single image, with the shape placed in a canonical frame of reference, i.e. disentangled from the pose.
Together, the model plus its inference network resemble a variational autoencoder~\citep{kingma14iclr} on pixels.
It represents 3D shapes in a compact latent embedding space, and has extra layers in the decoder corresponding to the mesh representation and renderer.
As we do not provide 3D supervision, the encoder and decoder must bootstrap and guide one another during training. The decoder learns the manifold of shapes, while at the same time the encoder learns to map images onto this.
This learning process is driven purely by the objective of reconstructing the training images.
While this is an ambiguous task and the model cannot guarantee to reconstruct the true shape of an object from a single image, its generative capability means that it always produces a plausible instance of the relevant class; the encoder ensures that this is consistent with the observed image.
This works because the generative model must learn to produce shapes that reproject well over {\em all} training images, starting from low-dimensional latent representations.
This creates an inductive bias towards regularity, which avoids degenerate solutions with unrealistic shapes that could, in isolation, explain each individual training image.

In \sect{experiments}, we demonstrate our method on 13 diverse object classes.
This includes several highly concave classes, which methods relying on silhouettes cannot learn correctly \citep{yan16nips,gadelha173dv,tulsiani17cvpr,tulsiani18cvpr-mvc}.
We first display samples from the distribution of shapes learnt by our model, showing that (i) the use of meshes yields smoother, more natural samples than those from voxel-based methods~\citep{gadelha173dv}, (ii) different mesh parameterisations are better suited to different object classes, and (iii) our samples are diverse and realistic, covering multiple modes of the training distribution.
We also demonstrate that our model learns a meaningful latent space, by showing that interpolating between points in it yields realistic intermediate samples.
%
%
We then quantitatively evaluate performance of our method on single-view reconstruction and pose estimation, showing that:
(i) it learns to predict pose, and disentangle it from shape, without either being given as supervision;
(ii) exploiting information from shading improves results over using silhouettes in the reconstruction loss, even when the model must learn to estimate the lighting parameters and disentangle them from surface normals;
(iii) when using a standard single white light, our model outperforms state-of-the-art 2D-supervised methods~\citep{kato18cvpr}, both with and without pose supervision, thanks to exploiting shading cues;
(iv) performance improves further when using multiple coloured lights, even approaching that of state-of-the-art 3D-supervised methods~\citep{fan17cvpr,richter18cvpr}.
Finally, we evaluate the impact of design choices such as different mesh parameterisations and latent space dimensionalities, showing which choices work well for different object classes.
\end{sloppypar}

A preliminary version of this work appeared as \citet{henderson18bmvc}.
That earlier version assumed fixed, known lighting parameters rather than explicitly reasoning over them; also here we present a much more extensive experimental evaluation.

\section{Related Work}
\label{sec:relwork}

\subsection{Learning single-image 3D reconstruction}
\label{sec:relwork/recon}

In the last three years, there has been a surge of interest in single-image 3D reconstruction; this has been enabled both by the growing maturity of deep learning techniques, and by the availability of large datasets of 3D shapes~\citep{shapenet15arxiv,wu15cvpr-shapenets}.
Among such methods, we differentiate between those requiring full 3D supervision (i.e. 3D shapes paired with images), and those that need only weaker 2D supervision (e.g. pose annotations);
our work here falls into the second category.

\paragraph{3D-supervised methods.}
\Citet{choy16eccv} apply a CNN to the input image, then pass the resulting features to a 3D deconvolutional network, that maps them to to occupancies of a $32^3$ voxel grid.
\Citet{girdhar16eccv} and \citet{wu16nips} proceed similarly, but pre-train a model to auto-encode or generate 3D shapes respectively, and regress images to the latent features of this model.
Instead of directly producing voxels, \citet{soltani17cvpr}, \citet{shin18cvpr} and \citet{richter18cvpr} output multiple depth-maps and/or silhouettes, from known (fixed) viewpoints; these are subsequently fused if a voxel reconstruction is required.
\Citet{fan17cvpr} and \citet{mandikal18bmvc} generate point clouds as the output, with networks and losses specialised to their order invariant structure.
Like ours, the concurrent works of \citet{groueix18cvpr} and \citet{wang18eccv} predict meshes, but parameterise them differently to us.
\Citet{tulsiani17cvpr-blocks} and \citet{niu18cvpr} both learn to map images to sets of cuboidal primitives, of fixed and variable cardinality respectively.
Finally, \citet{gwak173dv} and \citet{zhu17iccv-rethinking} present methods with slightly weaker requirements on ground-truth. As in the previous works, they require large numbers of 3D shapes and images; however, these do not need to be paired with each other. Instead, the images are annotated only with silhouettes.

\paragraph{2D-supervised methods.}
A few recent learning-based reconstruction techniques do not rely on 3D ground-truth; these are the closest in spirit to our own.
They typically work by passing input images through a CNN, which predicts a 3D representation, which is then rendered to form a reconstructed 2D silhouette; the loss is defined to minimise the difference between the reconstructed and original silhouettes.
This reliance on silhouettes means they cannot exploit shading and cannot learn to reconstruct concave object classes---in contrast to our approach.
Moreover, all these methods require stronger supervision than our own---they must be trained with ground-truth pose or keypoint annotations, and/or multiple views of each instance presented together during training.

\citet{rezende16nips} briefly discuss single-image reconstruction using a conditional generative model over meshes.
This models radial offsets to vertices of a spherical base mesh, conditioning on an input image.
The model is trained in a variational framework to maximise the reconstructed pixel likelihood.
It is demonstrated only on simple shapes such as cubes and cylinders.

\citet{yan16nips} present a method that takes single image as input, and yields a voxel reconstruction.
This is trained to predict voxels that reproject correctly to the input pixels, assuming the object poses for the training images are known.
The voxels are projected by computing a max operation along rays cast from each pixel into the voxel grid, at poses matching the input images.
The training objective is then to maximise the intersection-over-union (IOU) between these projected silhouettes and the silhouettes of the original images.
\citet{kato18cvpr} present a very similar method, but using meshes instead of voxels as the output representation.
It is again trained using the silhouette IOU as the loss, but also adds a smoothness regularisation term, penalising sharply creased edges.
\Citet{wiles17bmvc} propose a method that takes silhouette images as input, and produces rotated silhouettes as output; the input and output poses are provided.
To generate the rotated silhouettes, they predict voxels in 3D space, and project them by a max operation along rays.

\Citet{tulsiani17cvpr} also regress a voxel grid from a single image; however, the values in this voxel grid are treated as occupancy probabilities, which allows use of probabilistic ray termination~\citep{broadhurst01iccv} to enforce consistency with a silhouette or depth map.
Two concurrent works to ours, \citet{tulsiani18cvpr-mvc} and \citet{insafutdinov18nips}, extend this approach to the case where pose is not given at training time.
To disentangle shape and pose, they require that multiple views of each object instance be presented together during training; the model is then trained to reconstruct the silhouette for each view using its own predicted pose, but the shape predicted from some other view.
\Citet{yang18eccv} use the same principle to disentangle shape and pose, but assume that a small number of images are annotated with poses, which improves the accuracy significantly.

\Citet{vicente14cvpr} jointly reconstruct thousands of object instances in the PASCAL VOC 2012 dataset using keypoint and silhouette annotations, but without learning a model that can be applied to unseen images.
\Citet{kar15cvpr} train a CNN to predict keypoints, pose, and silhouette from an input image, and then optimise the parameters of a deformable model to fit the resulting estimates.
Concurrently with our work, \citet{kanazawa18eccv} present a method that takes a single image as input, and produces a textured 3D mesh as output.
The mesh is parameterised by offsets to the vertices of a learnt mean shape.
These three methods all require silhouette and keypoint annotations on the training images,
but only a single view of each instance.

\citet{novotny17iccv} learn to perform single-image reconstruction using videos as supervision.
Classical multi-view stereo methods are used to reconstruct the object instance in each video, and the reconstructions are used as ground-truth to train a regression model mapping images to 3D shapes.
%

\subsection{Generative models of 3D shape}
\label{sec:relwork/gen}

The last three years have also seen increasing interest in deep generative models of 3D shapes.
Again, these must typically be trained using large datasets of 3D shapes, while just one work requires only images~\citep{gadelha173dv}.

\paragraph{3D-supervised methods.}
\Citet{wu15cvpr-shapenets} and \citet{xie18cvpr} train deep energy-based models on voxel grids; \citet{huang15cgf} train one on surface points of 3D shapes, jointly with a decomposition into parts.
\Citet{wu16nips} and \citet{zhu18nips} present generative adversarial networks (GANs; \citealp{goodfellow14nips}) that directly model voxels using 3D convolutions; \citet{zhu18nips} also fine-tune theirs using 2D renderings.
\Citet{rezende16nips} and \citet{balashova183dv} both describe models of voxels, based on the variational autoencoder (VAE; \citealp{kingma14iclr}).
\Citet{nash17cgf} and \citet{gadelha18eccv} model point clouds, using different VAE-based formulations.
\Citet{achlioptas18icml} train an autoencoder for dimensionality reduction of point clouds, then a GAN on its embeddings.
\Citet{li17tog} and \citet{zou17iccv} model shapes as assembled from cuboidal primitives; \citet{li17tog} also add detail by modelling voxels within each primitive.
\Citet{tan18cvpr} present a VAE over parameters of meshes.
Calculating the actual vertex locations from these parameters requires a further energy-based optimisation, separate to their model.
Their method is not directly applicable to datasets with varying mesh topology, including ShapeNet and ModelNet.

\begin{sloppypar}
\paragraph{2D-supervised methods.}
\Citet{soltani17cvpr} train a VAE over groups of silhouettes from a set of known viewpoints; these may be fused to give a true 3D shape as a post-processing stage, separate to the probabilistic model.
 The only prior work that learns a true generative model of 3D shapes given just 2D images is \citet{gadelha173dv}; this is therefore the most similar in spirit to our own.
They use a GAN over voxels; these are projected to images by a simple max operation along rays, to give silhouettes.
A discriminator network ensures that projections of sampled voxels are indistinguishable from projections of ground-truth data.
This method does not require pose annotations, but they restrict poses to a set of just eight predefined viewpoints.
In contrast to our work, this method cannot learn concave shapes, due to its reliance on silhouettes.
Moreover, like other voxel-based methods, it cannot output smooth, arbitrarily-oriented surfaces.
\Citet{yang18eccv} apply this model as a prior for single-image reconstruction, but they require multiple views per instance during training.
\end{sloppypar}

\section{Generative Model}
\label{sec:generative}

Our goal is to build a probabilistic generative model of 3D meshes for a given object class.
For this to be trainable with 2D supervision, we cast the entire image-formation process as a directed model (\fig{flow}).
We assume that the content of an image can be explained by three independent latent components---the shape of the mesh, its pose relative to the camera, and the lighting.
These are modelled by three low-dimensional random variables, $\mathbf{z}$, $\theta$, and $\lambda$ respectively. The joint distribution over these and the resulting pixels $\mathbf{x}$ factorises as
$
  P(\mathbf{x} ,\, \mathbf{z} ,\, \theta ,\, \lambda)
   =
  P(\mathbf{z}) P(\theta) P(\lambda) P(\mathbf{x} \given \mathbf{z} ,\, \theta ,\, \lambda)
$.

\begin{sloppypar}
Following \citet{gadelha173dv}, \citet{yan16nips}, \citet{tulsiani17cvpr}, and \citet{wiles17bmvc}, we assume that the pose $\theta$ is parameterised by just the azimuth angle, with $\theta \sim \text{Uniform}(-\pi, \pi)$ (\fig{theta-and-lambda}a, bottom).
The camera is then placed at fixed distance and elevation relative to the object.
We similarly take $\lambda$ to be a single azimuth angle with uniform distribution, which specifies how a predefined set of directional light sources are to be rotated around the origin  (\fig{theta-and-lambda}a, top).
The number of lights, their colours, elevations, and relative azimuths are kept fixed.
We are free to choose these; our experiments include tri-directional coloured lighting, and a single white directional light source plus an ambient component.

Following recent works on deep latent variable models~\citep{kingma14iclr,goodfellow14nips}, we assume that the embedding vector $\mathbf{z}$ is drawn from a standard isotropic Gaussian, and then transformed by a deterministic \textit{decoder network},
 $F_{\phi}$, parameterised by weights $\phi$ which are to be learnt (Appendix~\ref{app:net-arch} details the architecture of this network).
This produces the \textit{mesh parameters} $\Pi = F_{\phi}(\mathbf{z})$.
Intuitively, the decoder network $F_{\phi}$ transforms and entangles the dimensions of $\mathbf{z}$ such that all values in the latent space map to plausible values for $\Pi$, even if these lie on a highly nonlinear manifold.
Note that our approach contrasts with previous models that directly output pixels~\citep{kingma14iclr,goodfellow14nips} or voxels~\citep{wu16nips,gadelha173dv,zhu18nips,balashova183dv} from a decoder network.

We use $\Pi$ as inputs to a fixed mesh parameterisation function $M(\Pi)$, which yields vertices $\mathbf{v}_{\text{object}}$ of triangles defining the shape of the object in 3D space, in a canonical pose (different options for $M$ are described below).
The vertices are transformed into camera space according to the pose $\theta$, by a fixed function $T$: $\mathbf{v}_{\text{camera}} = T(\mathbf{v}_{\text{object}},\, \theta)$.
They are then rendered into an RGB image $I_0 = \mathcal{G}(\mathbf{v}_{\text{camera}} ,\, \lambda)$ by a rasteriser $\mathcal{G}$ using Gouraud shading~\citep{gouraud71tc} and Lambertian surface reflectance~\citep{lambert60bk}.
\end{sloppypar}

The final observed pixel values $\mathbf{x}$ are modelled as independent Gaussian random variables, with means equal to the values in an $L$-level Gaussian pyramid~\citep{burt83tc}, whose base level equals $I_0$, and whose $L$\textsuperscript{th} level has smallest dimension equal to one:
\begin{equation}
    P_{\phi}(\mathbf{x} \given \mathbf{z},\, \theta ,\, \lambda) = \prod_l P_{\phi}(\mathbf{x}_l \given \mathbf{z} ,\, \theta ,\, \lambda)
\end{equation}
\vspace{-14pt}
\begin{equation}
    \mathbf{x}_l \sim \text{Normal}\left( I_l,\, \tfrac{\epsilon}{2^l} \right)
\end{equation}
\vspace{-12pt}
\begin{equation}
    I_0 = \mathcal{G}(T(M(F_{\phi}(\mathbf{z})),\, \theta),\, \lambda)
\end{equation}
\vspace{-12pt}
\begin{equation}
    I_{l+1} = I_l * k_G
\end{equation}

\noindent where $l$ indexes pyramid levels, $k_G$ is a small Gaussian kernel, $\epsilon$ is the noise magnitude at the base scale, and $*$ denotes convolution with stride two.
We use a multi-scale pyramid instead of just the raw pixel values to ensure that, during training, there will be gradient forces over long distances in the image, thus avoiding bad local minima where the reconstruction is far from the input.

\begin{figure}
  \includegraphics[width=\linewidth]{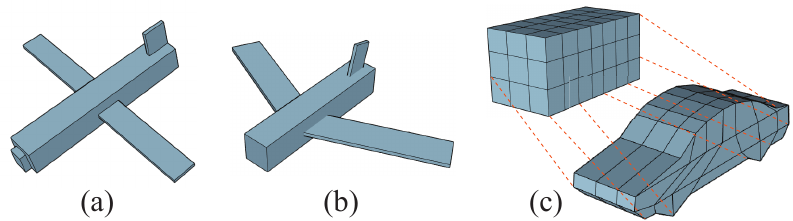}
  \caption{
  Mesh parameterisations: \textbf{ortho-block} \& \textbf{full-block} (assembly from cuboidal primitives, of fixed or varying orientation) are suited to objects consisting of compact parts (a-b); \textbf{subdivision} (per-vertex deformation of a subdivided cube) is suited to complex continuous surfaces (c).
  }
  \label{fig:mesh-parameterisations}
\end{figure}

\paragraph{Mesh parameterisations.}
After the decoder network has transformed the latent embedding $\mathbf{z}$ into the mesh parameters $\Pi$, these are converted to actual 3D vertices using a simple, non-learnt mesh-parameterisation function $M$.
One possible choice for $M$ is the identity function, in which case the decoder network directly outputs vertex locations.
However, initial experiments showed that this does not work well: it produces very irregular meshes with large numbers of intersecting triangles.
Conversely, using a more sophisticated form for $M$ enforces regularity of the mesh.
We use three different parameterisations in our experiments.

In our first parameterisation, $\Pi$ specifies the locations and scales of a fixed number of axis-aligned cuboidal \textit{primitives} (\fig{mesh-parameterisations}a), from which the mesh is assembled~\citep{zou17iccv,tulsiani17cvpr-blocks}.
Changing $\Pi$ can produce configurations with different topologies, depending which blocks touch or overlap, but all surfaces will always be axis-aligned.
The scale and location of each primitive are represented by 3D vectors, resulting in a total of six parameters per primitive.
In our experiments we call this \textbf{ortho-block}.

Our second parameterisation is strictly more powerful than the first: we still assemble the mesh from cuboidal primitives, but now associate each with a rotation, in addition to its location and scale.
Each rotation is parameterised as three Euler angles, yielding a total of nine parameters per primitive.
In our experiments we call this \textbf{full-block} (\fig{mesh-parameterisations}b).

The above parameterisations are naturally suited to objects composed of compact parts, but cannot represent complex continuous surfaces.
For these, we define a third parameterisation, \textbf{subdivision} (\fig{mesh-parameterisations}c).
This parameterisation is based on a single cuboid, centred at the origin; the edges and faces of the cuboid are subdivided several times along each axis.
Then, $\Pi$ specifies a list of 3D displacements, one per vertex, which deform the subdivided cube into the required shape.
In practice, we subdivide each edge into four segments, resulting in 98 vertices, hence 294 parameters.

\section{Variational Training}
\label{sec:training}

We wish to learn the parameters of our model from a training set of 2D images of objects of a single class.
More precisely, we assume access to a set of images $\{\mathbf{x}^{(i)}\}$, each showing an object with unknown shape, at an unknown pose, under unknown lighting.
Note that we do \textit{not} require that there are multiple views of each object (in contrast with \citet{yan16nips} and \citet{tulsiani18cvpr-mvc}), nor that the object poses are given as supervision (in contrast with \citet{yan16nips}, \citet{tulsiani17cvpr}, \citet{wiles17bmvc}, and \citet{kato18cvpr}).

We seek to maximise the marginal log-likelihood of the training set, which is given by $\sum_i \log P_{\phi}(\mathbf{x}^{(i)})$, with respect to $\phi$.
For each image, we have
\begingroup\makeatletter\def\f@size{9}\check@mathfonts
\begin{equation}
\label{eq:loglik}
\log P_{\phi}(\mathbf{x}^{(i)}) =
\log \int_{\mathbf{z},\theta,\lambda} P_{\phi}(\mathbf{x}^{(i)} \given \mathbf{z}, \theta, \lambda) P(\mathbf{z}) P(\theta) P(\lambda) \, \mathrm{d}\mathbf{z} \, \mathrm{d}\theta \, \mathrm{d}\lambda
\end{equation}\endgroup
Unfortunately this is intractable, due to the integral over the latent variables $\mathbf{z}$ (shape), $\theta$ (pose), and $\lambda$ (lighting).
Hence, we use amortised variational inference, in the form of stochastic gradient variational Bayes \citep{kingma14iclr,rezende14icml}.
This introduces an approximate posterior $Q_{\omega}(\mathbf{z}, \theta, \lambda \given \mathbf{x})$, parameterised by some $\omega$ that we learn jointly with the model parameters $\phi$.
Intuitively, $Q$ maps an image $\mathbf{x}$ to a distribution over likely values of the latent variables $\mathbf{z}$, $\theta$, and $\lambda$.
Instead of the log-likelihood \eqn{loglik}, we then maximise the \textit{evidence lower bound} (ELBO):
\begin{multline}
\label{eq:elbo}
  \mathop{\mathbb{E}}_{\mathbf{z},\, \theta,\, \lambda \sim Q_{\omega}(\mathbf{z},\, \theta,\, \lambda \given \mathbf{x}^{(i)})}\left[
    \log P_{\phi}( \mathbf{x}^{(i)} \given \mathbf{z},\, \theta,\, \lambda )
  \right] \\
  - \kldiv{
    Q_{\omega}(\mathbf{z},\, \theta,\, \lambda \given \mathbf{x}^{(i)})
  }{
    P(\mathbf{z}) P(\theta) P(\lambda)
  }
  \le
  \log P_{\phi}(\mathbf{x}^{(i)})
\end{multline}
This lower-bound on the log-likelihood can be evaluated efficiently, as the necessary expectation is now with respect to $Q$, for which we are free to choose a tractable form.
The expectation can then be approximated using a single sample.

\begin{figure}
  \centering
  \includegraphics[width=0.46\linewidth]{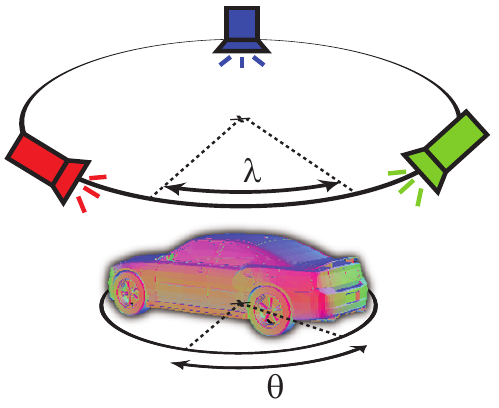}
  \includegraphics[width=0.52\linewidth]{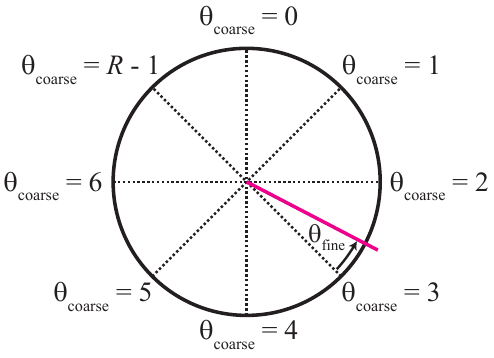} \\
  \textbf{(a)} \hspace{3.5cm} \textbf{(b)}
  \caption{
  \textbf{(a)} We parameterise the object pose relative to the camera by the azimuth angle $\theta$, and rotate the lights around the object as a group according to a second azimuth angle $\lambda$.
  \textbf{(b)} To avoid degenerate solutions, we discretise $\theta$ into coarse and fine components, with $\theta_{\mathrm{coarse}}$ categorically distributed over $R$ bins, and $\theta_{\mathrm{fine}}$ specifying a small offset relative to this.
  For example, to represent the azimuth indicated by the pink line, $\theta_{\mathrm{coarse}} = 3$ and $\theta_{\mathrm{fine}} = -18^\circ$.
  The encoder network outputs softmax logits $\mathbf{\rho}$ for a categorical variational distribution over $\theta_{\mathrm{coarse}}$, and the mean $\xi$ and standard deviation $\zeta$ of a Gaussian variational distribution over $\theta_{\mathrm{fine}}$, with $\xi$ bounded to the range $( -\pi / R ,\, \pi / R )$.
  }
  \label{fig:theta-and-lambda}
\end{figure}

We let $Q$ be a mean-field approximation, i.e. given by a product of independent variational distributions:
\begin{equation}
Q_{\omega}(\mathbf{z}, \theta, \lambda \given \mathbf{x}) =
Q_{\omega}(\mathbf{z} \given \mathbf{x})
Q_{\omega}(\theta \given \mathbf{x})
Q_{\omega}(\lambda \given \mathbf{x})
\end{equation}
The parameters of these distributions are produced by an \textit{encoder network}, $\mathrm{enc}_{\omega}(\mathbf{x})$, which takes the image $\mathbf{x}$ as input.
For this encoder network we use a small CNN with architecture similar to \citet{wiles17bmvc} (see Appendix~\ref{app:net-arch}).
We now describe the form of the variational distribution for each of the variables $\mathbf{z}$, $\theta$, and $\lambda$.

\paragraph{Shape.}
For the shape embedding $\mathbf{z}$, the variational posterior distribution $Q_{\omega}(\mathbf{z} \given \mathbf{x})$  is a multivariate Gaussian with diagonal covariance.
The mean and variance of each latent dimension are produced by the encoder network.
When training with multiple views per instance, we apply the encoder network to each image separately, then calculate the final shape embedding $\mathbf{z}$ by max-pooling each dimension over all views.

\begin{sloppypar}
\paragraph{Pose.}
For the pose $\theta$, we could similarly use a Gaussian posterior.
However, many objects are roughly symmetric with respect to rotation, and so the true posterior is typically multi-modal.
We capture this multi-modality by decomposing the rotation into coarse and fine parts~\citep{mousavian17cvpr}: an integer random variable $\theta_{\text{coarse}}$ that chooses from $R_\theta$ rotation bins, and a small Gaussian offset $\theta_{\text{fine}}$ relative to this (\fig{theta-and-lambda}b):
\begin{equation}
  \theta = -\pi + \theta_{\text{coarse}} \frac{2\pi}{R_\theta} + \theta_{\text{fine}}
\end{equation}
We apply this transformation in both the generative $P(\theta)$ and variational $Q_{\omega}(\theta)$, giving
\begin{equation}
  \label{eq:theta-coarse-prior}
  P(\theta_{\text{coarse}} = r) = 1/R_\theta
\end{equation}
\begin{equation}
  P(\theta_{\text{fine}}) = \text{Normal}(\theta_{\text{fine}} \given 0,\, \pi / R_\theta )
\end{equation}
\begin{equation}
  Q_{\omega}\left( \theta_{\text{coarse}} = r \Given \mathbf{x}^{(i)} \right) = \rho_r^\theta \left( \mathbf{x}^{(i)} \right)
\end{equation}
\begin{equation}
  Q_{\omega}(\theta_{\text{fine}}) = \text{Normal}\left( \theta_{\text{fine}} \Given \xi^\theta(\mathbf{x}^{(i)}),\, \zeta^\theta(\mathbf{x}^{(i)}) \right)
\end{equation}
where the variational parameters $\rho_r^\theta, \xi^\theta, \zeta^\theta$ for image $\mathbf{x}^{(i)}$ are again estimated by the encoder network $\mathrm{enc}_{\omega}(\mathbf{x}^{(i)})$.
Specifically, the encoder uses a softmax output to parameterise $\mathbf{\rho}^\theta$, and restricts $\xi^\theta$ to lie in the range $( -\pi / R_\theta ,\, \pi / R_\theta )$, ensuring that the fine rotation is indeed a small perturbation, so the model must correctly use it in conjunction with $\theta_\mathrm{coarse}$.
\end{sloppypar}

Provided $R_\theta$ is sufficiently small, we can integrate directly with respect to $\theta_{\text{coarse}}$ when evaluating \eqn{elbo}, i.e. sum over all possible rotations.
%
While this allows our training process to reason over different poses, it is still prone to predicting the same pose $\theta$ for every image;
clearly this does not correspond to the prior on $\theta$ given by \eqn{theta-coarse-prior}.
The model is therefore relying on the shape embedding $\mathbf{z}$ to model all variability, rather than disentangling shape and pose.
The ELBO \eqn{elbo} does include a KL-divergence term that should encourage latent variables to match their prior.
However, it does not have a useful effect for $\theta_{\text{coarse}}$: minimising the KL divergence from a uniform distribution for each sample individually corresponds to independently minimising all the probabilities $Q_{\omega}(\theta_{\text{coarse}})$, which does not encourage uniformity of the full distribution.
The effect we desire is to match the aggregated posterior distribution $\left\langle Q_{\omega}(\theta \given \mathbf{x}^{(i)}) \right\rangle_i$ to the prior $P(\theta)$, where $\langle \,\cdot\, \rangle_i$ is the empirical mean over the training set.
As $\theta_{\text{coarse}}$ follows a categorical distribution in both generative and variational models, we can directly minimise the L1 distance between the aggregated posterior and the prior
\begingroup\makeatletter\def\f@size{8}\check@mathfonts
\begin{equation}
  \sum_r^{R_\theta} \bigg\lvert
  \left\langle
  Q_{\omega}\left(\theta_{\text{coarse}} = r \given \mathbf{x}^{(i)}\right)
  \right\rangle_i
  - P\left(\theta_{\text{coarse}} = r\right)
  \bigg\rvert
  =
  \sum_r^{R_\theta} \bigg\lvert
  \left\langle
  \rho_r^\theta(\mathbf{x}^{(i)})
  \right\rangle_i
  - \frac{1}{R_\theta} \;
  \bigg\rvert
\end{equation}\endgroup
We use this term in place of $\kldiv{ Q(\theta_{\text{coarse}} \given \mathbf{x}^{(i)}) }{ P(\theta_{\text{coarse}}) }$ in our loss, approximating the empirical mean with a single minibatch.

\paragraph{Lighting.}
For the lighting angle $\lambda$, we perform the same decomposition into coarse and fine components as for $\theta$, giving new variables $\lambda_\mathrm{coarse}$ and $\lambda_\mathrm{fine}$, with $\lambda_\mathrm{coarse}$ selecting from among $R_\lambda$ bins.
Analogously to pose, $\lambda_\mathrm{coarse}$ has a categorical variational distribution parameterised by a softmax output $\rho^\lambda$ from the encoder, and $\lambda_\mathrm{fine}$ has a Gaussian variational distribution with parameters $\xi^\lambda$ and $\zeta^\lambda$.
Again, we integrate over $\lambda_\mathrm{coarse}$, so the training process reasons over many possible lighting angles for each image, increasing the predicted probability of the one giving the best reconstruction.
We also regularise the aggregated posterior distribution of $\lambda_\mathrm{coarse}$ towards a uniform distribution.
%

\paragraph{Loss.}
Our final loss function for a minibatch $\mathcal{B}$ is then given by

\begingroup\makeatletter\def\f@size{8}\check@mathfonts
\begin{equation}
\label{eq:loss}
\begin{split}
  \sum_{r_\theta}^{R_\theta} \;
  \sum_{r_\lambda}^{R_\lambda}
  \Bigg\{
    - & \bigg\langle
      \mathop{\mathbb{E}}_{
        \mathbf{z},\, \theta_{\text{fine}},\, \lambda_{\text{fine}} \sim Q_{\omega}
      }\bigg[
\\
        & \qquad \log P_{\phi}\!\left( \mathbf{x}^{(i)} \Given \mathbf{z},\, \theta_{\text{coarse}} = r_\theta,\, \theta_{\text{fine}},\, \lambda_{\text{coarse}} = r_\lambda,\, \lambda_{\text{fine}} \right)
  \\
      & 
  \bigg]
  \;
  \rho^\theta_{r_\theta} \left(\mathbf{x}^{(i)} \right) \,
  \rho^\lambda_{r_\lambda} \left(\mathbf{x}^{(i)} \right)
  \bigg\rangle_{\!i \in \mathcal{B}}
\Bigg\}
\\
+ & \, \alpha \;
\sum_r^{R_\theta}
\Bigg\{
  \bigg\lvert
  \!
  \left\langle
  \rho_r^\theta \left(\mathbf{x}^{(i)} \right)
  \right\rangle_{\!i \in \mathcal{B}}
  \!
  - \frac{1}{R_\theta} \;
  \bigg\rvert
\Bigg\}
  \\
+ & \, \alpha \;
\sum_r^{R_\lambda}
\Bigg\{
  \bigg\lvert
  \!
  \left\langle
  \rho_r^\lambda \left(\mathbf{x}^{(i)} \right)
  \right\rangle_{\!i \in \mathcal{B}}
  \!
  - \frac{1}{R_\lambda} \;
  \bigg\rvert
\Bigg\}
  \\
  + & \, \beta \;
  \bigg\langle
    \kldiv{
      Q_{\omega}\left( \mathbf{z}, \theta_{\text{fine}}, \lambda_{\text{fine}} \Given \mathbf{x}^{(i)} \right)
    }{
      P(\mathbf{z}) P(\theta_{\text{fine}}) P(\lambda_{\text{fine}})
    }
  \bigg\rangle_{\!i \in \mathcal{B}}
\end{split}
\end{equation}
\endgroup
where $\beta$ increases the relative weight of the KL term as in \citet{higgins17iclr}, and $\alpha$ controls the strength of the prior-matching terms for pose and lighting.
We minimise \eqn{loss} with respect to $\phi$ and $\omega$ using ADAM~\citep{kingma15iclr} with gradient clipping, applying the reparameterisation trick to handle the Gaussian random variables~\citep{kingma14iclr,rezende14icml}.
%
Hyperparameters are given in Appendix~\ref{app:hyperparams}.

\paragraph{Differentiable rendering.}
Note that optimising \eqn{loss} by gradient descent requires differentiating through the mesh-rendering operation $\mathcal{G}$ used to calculate $P_{\phi}(\mathbf{x} \given \mathbf{z} ,\, \theta ,\, \lambda)$, to find the derivative of the pixels with respect to the vertex locations and colours.
While computing exact derivatives of $\mathcal{G}$ is very expensive, \citet{loper14eccv} describe an efficient approximation.
We employ a similar technique here, and have made our TensorFlow implementation publicly available\footnote{\textit{DIRT: a fast Differentiable Renderer for TensorFlow}, available at \url{https://github.com/pmh47/dirt}}.

\section{Experiments}
\label{sec:experiments}

%
We follow recent works \citep{gadelha173dv,yan16nips,tulsiani17cvpr,fan17cvpr,kato18cvpr,tulsiani18cvpr-mvc,richter18cvpr,yang18eccv} and evaluate our approach using the ShapeNet dataset~\citep{shapenet15arxiv}.
Using synthetic data has two advantages: it allows controlled experiments modifying lighting and other parameters, and it lets us evaluate the reconstruction accuracy using the ground-truth 3D shapes.

We begin by demonstrating that our method successfully learns to generate and reconstruct 13 different object classes (\sect{experiments/diverse-classes}).
These include the top ten most frequent classes of ShapeNet, plus three others (\textit{bathtub}, \textit{jar}, and \textit{pot}) that we select because they are smooth and concave, meaning that prior methods using voxels and silhouettes cannot learn and represent them faithfully, as shading information is needed to handle them correctly.

\begin{sloppypar}
We then rigorously evaluate the performance of our model in different settings, focusing on four classes (\textit{aeroplane}, \textit{car}, \textit{chair}, and \textit{sofa}).
The first three are used in \citet{yan16nips}, \citet{tulsiani17cvpr}, \citet{kato18cvpr}, and \citet{tulsiani18cvpr-mvc}, while the fourth is a highly concave class that is hard to handle by silhouette-based approaches.
We conduct experiments varying the following factors:
\begin{itemize}

  \item
  \textbf{Mesh parameterisations} (\sect{experiments/paramns}): We evaluate the three parameterisations described in \sect{generative}: \textbf{ortho-block}, \textbf{full-block}, and \textbf{subdivision}.

  \item
  \textbf{Single white light vs. three coloured lights} (\sect{experiments/lighting}): Unlike previous works using silhouettes (\sect{relwork}), our method is able to exploit shading in the training images.
  We test in two settings:
  (i) illumination by three coloured directional lights (\textbf{colour}, \fig{lighting}a); and
  (ii) illumination by one white directional light plus a white ambient component (\textbf{white}, \fig{lighting}b).

  \item
  \textbf{Fixed vs. varying lighting} (\sect{experiments/lighting}):
  The variable $\lambda$ represents a rotation of all the lights together around the vertical axis (\sect{generative}).
  We conduct experiments in two settings:
  (i) $\lambda$ is kept fixed across all training and test images, and is known to the generative model (\textbf{fixed}); and
  (ii) $\lambda$ is chosen randomly for each training/test image, and is not provided to the model (\textbf{varying}).
  In the latter setting, the model must learn to disentangle the effects of lighting angle and surface orientation on the observed shading.

  \item
  \textbf{Silhouette vs. shading in the loss} (\sect{experiments/lighting}):
  We typically calculate the reconstruction loss (pixel log-likelihood) over the RGB shaded image (\textbf{shading}), but for comparison with 2D-supervised silhouette-based methods (\sect{relwork})
  we also experiment with using only the silhouette in the loss (\textbf{silhouette}), disregarding differences in shading between the input and reconstructed pixels.

  \item
  \textbf{Latent space dimensionality} (\sect{experiments/latent-space}): We experiment with different sizes for the latent shape embedding $\mathbf{z}$, which affects the representational power of our model.
  We found that 12 dimensions gave good results in initial experiments, and use this value for all experiments apart from \sect{experiments/latent-space}, where we evaluate its impact.

  \item
  \textbf{Multiple views} (\sect{experiments/multi-view}): \citet{yan16nips}, \citet{wiles17bmvc}, \citet{tulsiani18cvpr-mvc} and \citet{yang18eccv} require that multiple views of each instance are presented together in each training batch, and \citet{tulsiani17cvpr} also focus on this setting.
  Our model does not require this, but for comparison we include results with three views per instance at training time, and either one or three at test time.

  \item
  \textbf{Pose supervision:} Most previous works that train for 3D reconstruction with 2D supervision require the ground-truth pose of each training instance~\citep{yan16nips,wiles17bmvc,tulsiani17cvpr}.
  While our method does not need this, we evaluate whether it can benefit from it, in each of the settings described above (we report these results in their corresponding sections).

\end{itemize}
Finally, we compare the performance of our model to several prior and concurrent works on generation and reconstruction, using various degrees of supervision (\sect{experiments/prior-comparison}).
\end{sloppypar}

\paragraph{Evaluation metrics.}
We benchmark our reconstruction and pose estimation accuracy on a held-out test set, following the protocol of \citet{yan16nips}, where each object is presented at 24 different poses, and statistics are aggregated across objects and poses.
We use the following measures:
\begin{itemize}
  \item
  \textit{iou}: to measure the shape reconstruction error, we calculate the mean intersection-over-union between the predicted and ground-truth shapes. For this we voxelise both meshes at a resolution of $32^3$. This is the metric used by recent works on reconstruction with 2D supervision~\citep[e.g.][]{yan16nips,tulsiani17cvpr,kato18cvpr,wiles17bmvc}
  \item
  \textit{err}: to measure the pose estimation error, we calculate the median error in degrees of predicted rotations
  \item
  \textit{acc}: again to evaluate pose estimation, we measure the fraction of instances whose predicted rotation is within $30^\circ$ of the ground-truth rotation.
\end{itemize}
Note that the metrics \textit{err} and \textit{acc} are used by \citet{tulsiani18cvpr-mvc} to evaluate pose estimation in a similar setting to ours.

\paragraph{Training minibatches.}
%
Each ShapeNet mesh is randomly assigned to either the training set (80\% of meshes) or the test set.
During training, we construct each minibatch by randomly sampling 128 meshes from the relevant class, uniformly with replacement.
For each selected mesh, we render a single image, using a pose sampled from $\text{Uniform}(-\pi,\,\pi)$ (and also sampling a lighting angle for experiments with varying lighting).
Only these images are used to train the model, not the meshes themselves.
In experiments using multiple views, we instead sample 64 meshes and three poses per mesh, and correspondingly render three images.

\begin{table}[t]
  \centering
  \caption{
  Reconstruction and pose estimation performance for the ten most-frequent classes in ShapeNet (first ten rows), plus three smooth, concave classes that methods based on voxels and silhouettes cannot handle (last three rows).
  Metrics:
  \textit{iou} measures shape reconstruction accuracy when pose supervision is not given (1 = best, 0 = worst);
    \textit{err} and \textit{acc} measure pose estimation in this case, which requires the model to disentangle shape and pose (\textit{err}: best = 0, worst = 180; \textit{acc}: best = 1, worse = 0);
    \itiougt{} measures shape reconstruction accuracy when pose supervision is given during training (best = 1, worst = 0).
  Note that table, lamp, pot, and jar all typically have rotational symmetry, and as such, it is not possible to define an unambiguous reference frame; this results in high values for \textit{err} and low for \textit{acc}.
  Experimental setting: subdivision, single-view training, fixed colour lighting, shading loss.
  }
  \label{tab:recon-all-classes}

  \begin{tabular}{@{} lllll @{}}
    \toprule
    & iou & err & acc & \iougt \\
    & \textit{(shape)} & \textit{(pose)} & \textit{(pose)} & \textit{(shape)} \\
    \midrule
    table & 0.44 & 89.3 & 0.39 & 0.49 \\
    chair & 0.39 & 7.9 & 0.65 & 0.51 \\
    airplane & 0.55 & 1.4 & 0.90 & 0.59 \\
    car & 0.77 & 4.7 & 0.84 & 0.82 \\
    sofa & 0.59 & 6.5 & 0.88 & 0.71 \\
    rifle & 0.54 & 9.0 & 0.68 & 0.61 \\
    lamp & 0.40 & 87.7 & 0.19 & 0.41 \\
    vessel & 0.48 & 9.8 & 0.59 & 0.58 \\
    bench & 0.35 & 5.1 & 0.71 & 0.44 \\
    loudspeaker & 0.41 & 81.7 & 0.28 & 0.54 \\
    \midrule
    bathtub & 0.54 & 9.7 & 0.54 & 0.57 \\
    pot & 0.49 & 90.4 & 0.20 & 0.53  \\
    jar & 0.49 & 93.1 & 0.16 & 0.52  \\
    \bottomrule
  \end{tabular}
\end{table}

\begin{figure*}
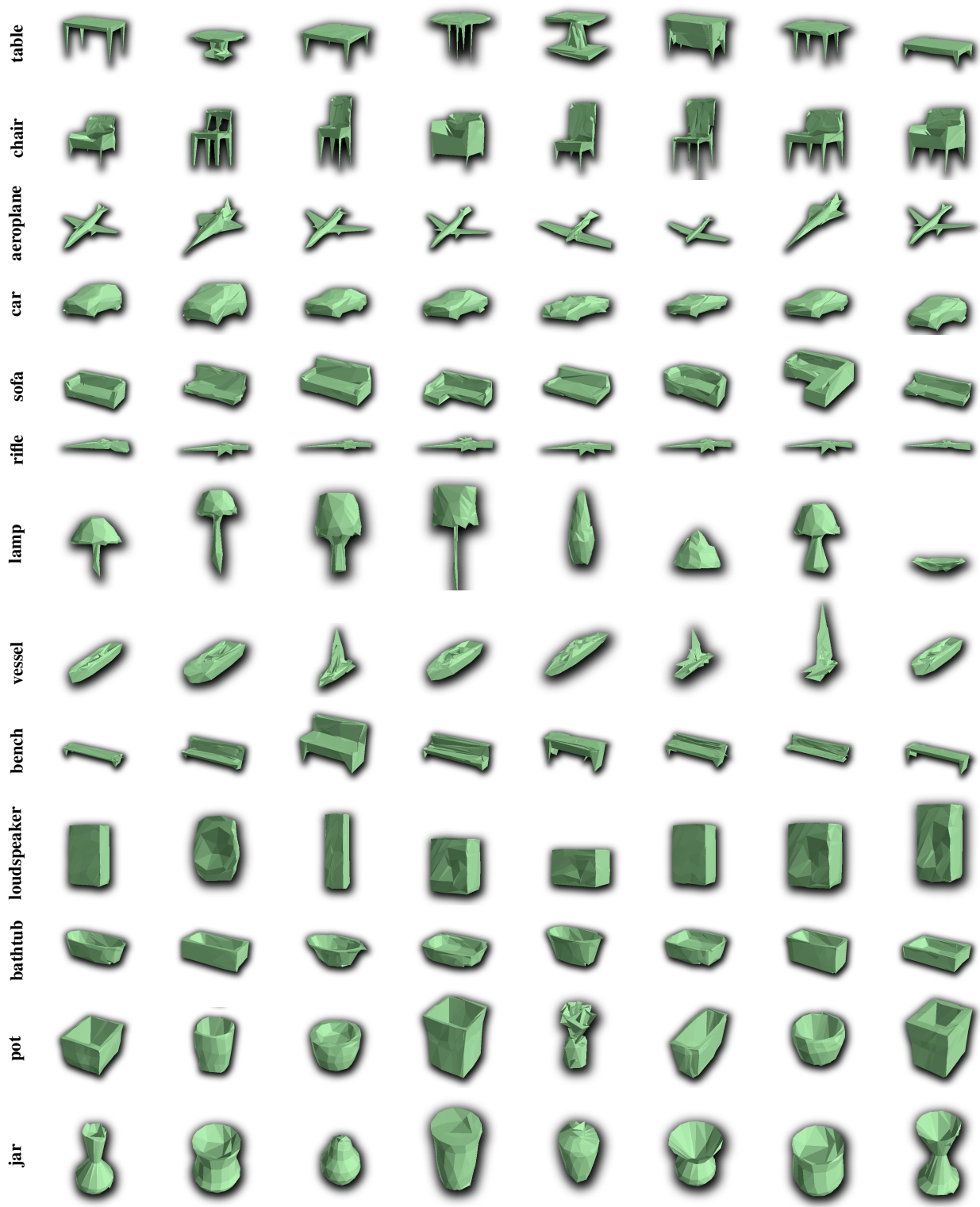

  \centering
  \newcommand{\classcol}[2]{
    \includegraphics[width=1.5cm]{images_#1_gen#2.png} 
  }
  \newcommand{\classrow}[2]{
    \rotatebox{90}{\hspace{#2}\textbf{#1}} &
    \classcol{#1}{1} & \classcol{#1}{2} & \classcol{#1}{3} & \classcol{#1}{4} &
    \classcol{#1}{5} & \classcol{#1}{6} & \classcol{#1}{7} & \classcol{#1}{8} 
    \\
  }
  \begin{tabular}{@{}c cccccccc @{}}
    \classrow{table}{0.7em}
    \classrow{chair}{1.2em}
    \classrow{aeroplane}{0em}
    \classrow{car}{1em}
    \classrow{sofa}{1em}
    \classrow{rifle}{0em}
    \classrow{lamp}{1.5em}
    \classrow{vessel}{0.7em}
    \classrow{bench}{0.5em}
    \classrow{loudspeaker}{0.3em}
    \classrow{bathtub}{0em}
    \classrow{pot}{1.5em}
    \classrow{jar}{2.5em}
  \end{tabular}
  \caption{
  Samples from our model for the ten most frequent classes in ShapeNet in order of decreasing frequency, plus three other interesting classes.
  Note the diversity and realism of our samples, which faithfully capture multimodal shape distributions, e.g. both straight and right-angled sofas, boats with and without sails, and straight- and delta-wing aeroplanes.
  We successfully learn models for the highly concave classes sofa, bathtub, pot, and jar, enabled by the fact that we exploit shading cues during training.
  Experimental setting: subdivision, fixed colour lighting, shading loss.
  }
  \label{fig:gen-all-classes}
\end{figure*}

\begin{figure*}
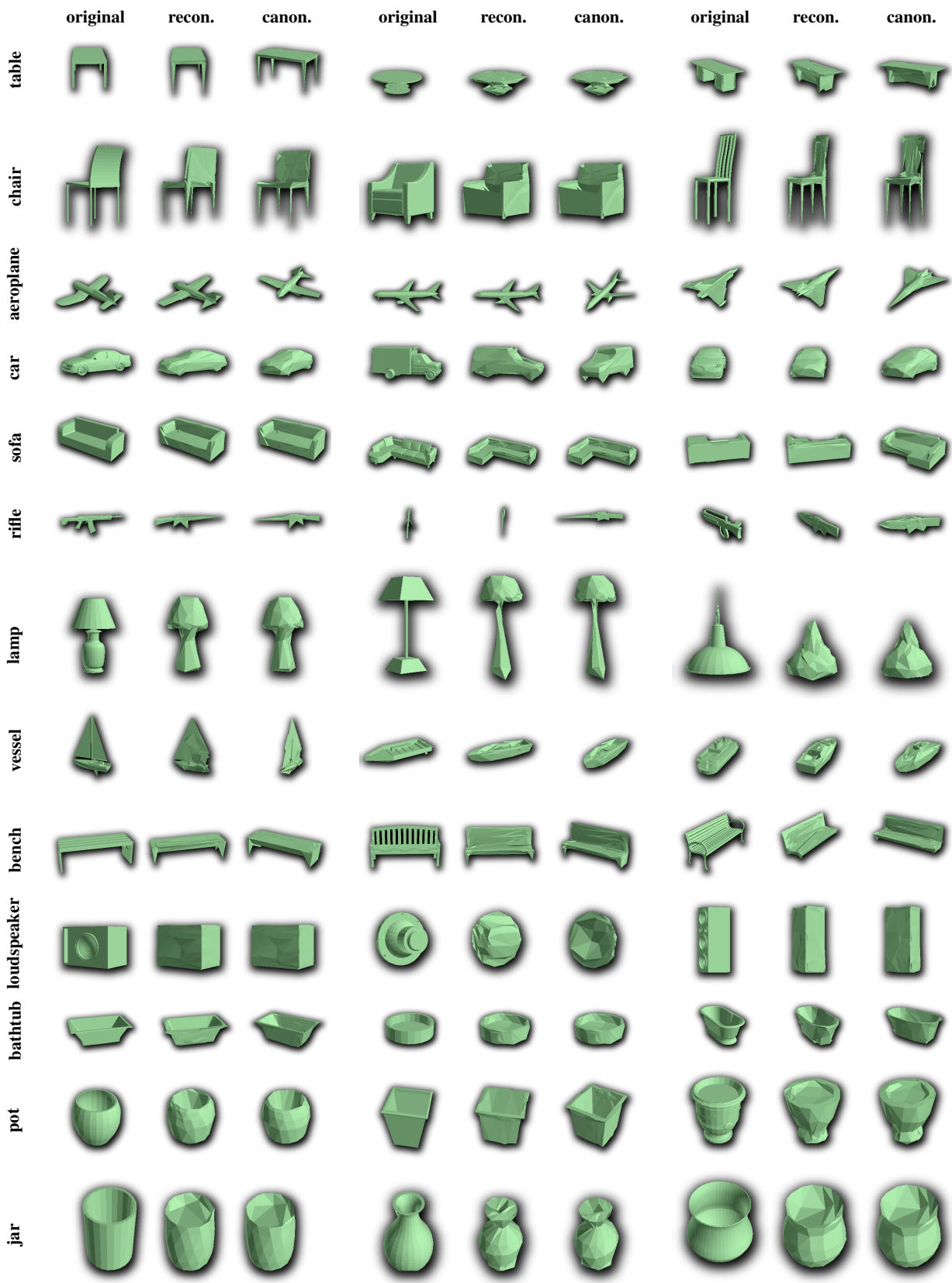

  \centering
  \newcommand{\classcol}[2]{
  \includegraphics[width=4.8cm]{images_#1_recon#2.png}
  }
  \newcommand{\classrow}[2]{
  \rotatebox{90}{\hspace{#2}\textbf{#1}} &
  \classcol{#1}{1} & \classcol{#1}{2} & \classcol{#1}{3} \\
  }
  \begin{tabular}{@{}c ccc @{}}
    &
    \begin{tabular}{>{\centering}m{1.2cm} >{\centering}m{1.2cm} >{\centering}m{1.2cm}} \textbf{original} & \textbf{recon.} & \textbf{canon.} \end{tabular} &
    \begin{tabular}{>{\centering}m{1.2cm} >{\centering}m{1.2cm} >{\centering}m{1.2cm}} \textbf{original} & \textbf{recon.} & \textbf{canon.} \end{tabular} &
    \begin{tabular}{>{\centering}m{1.2cm} >{\centering}m{1.2cm} >{\centering}m{1.2cm}} \textbf{original} & \textbf{recon.} & \textbf{canon.} \end{tabular} \\
    \classrow{table}{1.5em}
    \classrow{chair}{2em}
    \classrow{aeroplane}{0em}
    \classrow{car}{1em}
    \classrow{sofa}{1.2em}
    \classrow{rifle}{1em}
    \classrow{lamp}{2em}
    \classrow{vessel}{1.5em}
    \classrow{bench}{0.7em}
    \classrow{loudspeaker}{0em}
    \classrow{bathtub}{0em}
    \classrow{pot}{1.5em}
    \classrow{jar}{2.5em}
  \end{tabular}
  \caption{
  Qualitative examples of reconstructions for different object classes.  Each group of three images shows (i) ShapeNet ground-truth; (ii) our reconstruction; (iii) reconstruction placed in a canonical pose, with the different viewpoint revealing hidden parts of the shape. Experimental setting: subdivision, single-view training, fixed colour lighting, shading loss.}
  \label{fig:recon-all-classes}
\end{figure*}

\begin{figure*}
  \centering
  \newcommand{\classrow}[1]{
  \rotatebox{90}{\textbf{#1}} &
  \includegraphics[width=1.5cm]{images_#1_ortho_gen1} &
  \includegraphics[width=1.5cm]{images_#1_ortho_gen2} &
  \includegraphics[width=1.5cm]{images_#1_ortho_gen3} &
  \includegraphics[width=1.5cm]{images_#1_blocks_gen1} &
  \includegraphics[width=1.5cm]{images_#1_blocks_gen2} &
  \includegraphics[width=1.5cm]{images_#1_blocks_gen3} &
  \includegraphics[width=1.5cm]{images_#1_gen1} &
  \includegraphics[width=1.5cm]{images_#1_gen2} &
  \includegraphics[width=1.5cm]{images_#1_gen3} \\
  }
  \begin{tabular}{@{} c ccc | ccc | ccc @{}}
    & \multicolumn{3}{c|}{\textbf{ortho-block}} & \multicolumn{3}{c|}{\textbf{full-block}} & \multicolumn{3}{c}{\textbf{subdivision} } \\
    \classrow{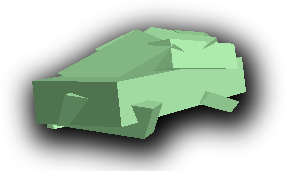}
    \classrow{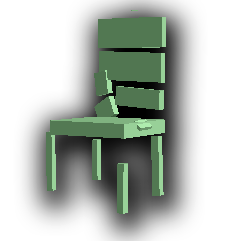}
    \classrow{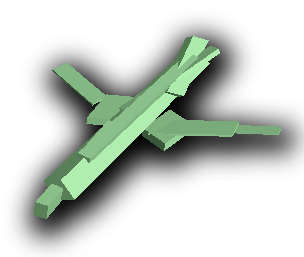}
    \classrow{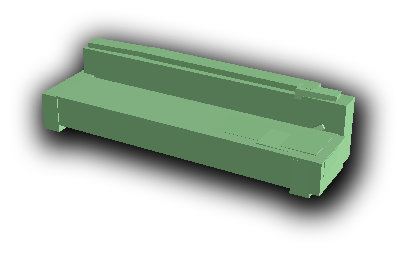}
  \end{tabular}
  \caption{Samples for four object classes, using our three different mesh parameterisations. \textbf{ortho-block} and \textbf{full-block} perform well for sofas and reasonably for chairs, but are less well-suited to aeroplanes and cars, which are naturally represented as smooth surfaces. \textbf{subdivision} gives good results for all four object classes.
  }
  \label{fig:gen-different-paramns}
\end{figure*}

\begin{figure*}
  \centering
  \includegraphics[width=\linewidth]{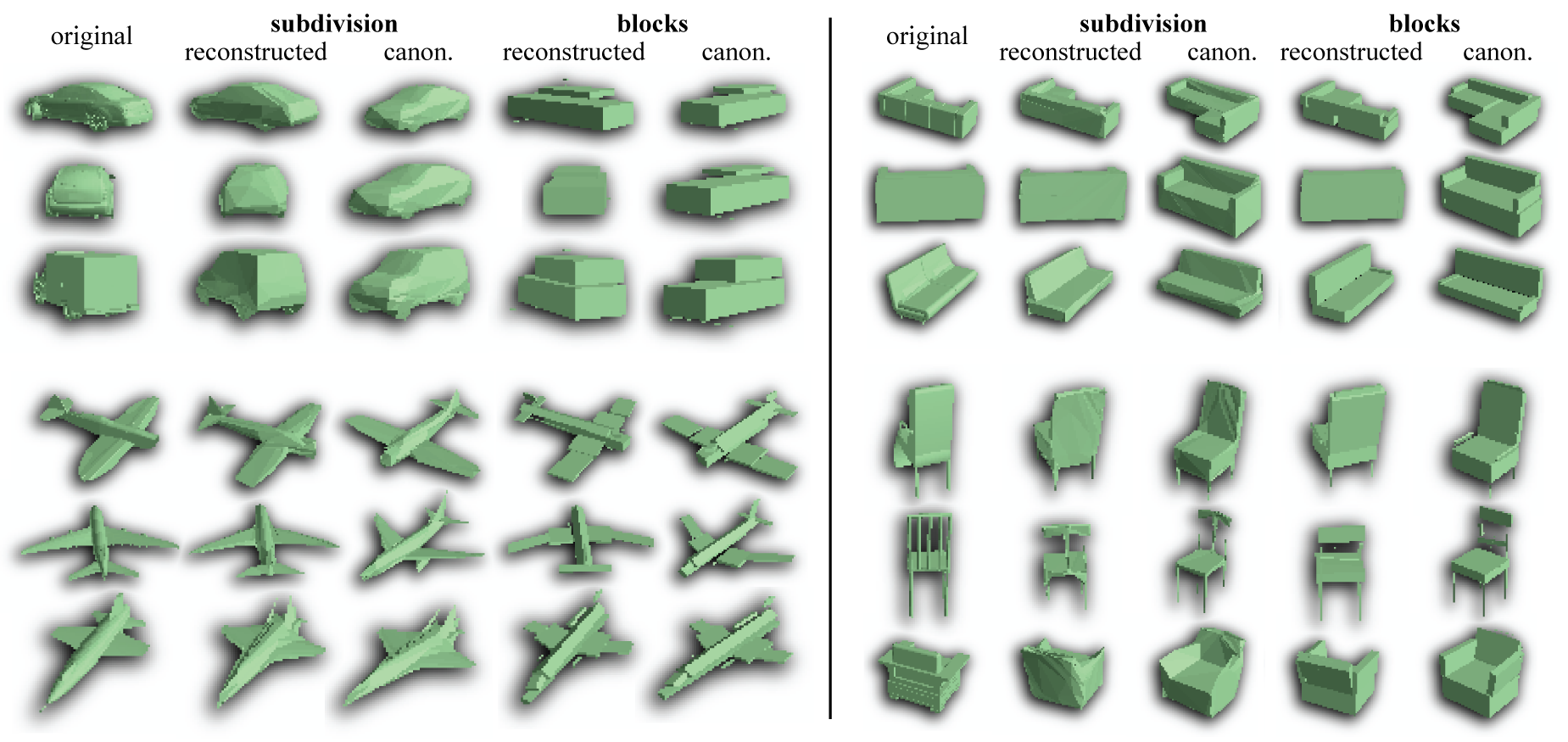}
  \caption{
  Qualitative examples of reconstructions, using different mesh parameterisations. Each row of five images shows (i) ShapeNet ground-truth; (ii) our reconstruction with \textbf{subdivision} parameterisation; (iii) reconstruction placed in a canonical pose; (iv) our reconstruction with \textbf{blocks}; (v) canonical-pose reconstruction. Experimental setting: single-view training, fixed colour lighting, shading loss.}
  \label{fig:recon-different-paramns}
\end{figure*}

\subsection{Generating and reconstructing diverse object classes}
\label{sec:experiments/diverse-classes}

We train a separate model for each of the 13 object classes mentioned above, using \textbf{subdivision} parameterisation.
Samples generated from these models are shown in \fig{gen-all-classes}.
%
We see that the sampled shapes are realistic, and the models have learnt a prior that encompasses the space of valid shapes for each class.
Moreover, the samples are diverse: the models generate various different styles for each class.
For example, for \textit{sofa}, both straight and right-angled (modular) designs are sampled;
for \textit{aeroplane}, both civilian airliners and military (delta-wing) styles are sampled;
for \textit{pot}, square, round, and elongated, forms are sampled; and, for \textit{vessel}, boats both with and without sails are sampled.
Note also that our samples incorporate smoothly curved surfaces (e.g. \textit{car}, \textit{jar}) and slanted edges (e.g. \textit{aeroplane}), which voxel-based methods cannot represent (\sect{experiments/prior-comparison} gives a detailed comparison with one such method~\citep{gadelha173dv}).


Reconstruction results are given in \tab{recon-all-classes}, with qualitative results in \fig{recon-all-classes}.
We use fixed colour lighting, shading loss, single-view training, and no pose supervision (columns \textit{iou, err, acc}); we also report \textit{iou} when using pose supervision in column \itiougt.
We see that the highest reconstruction accuracy (\textit{iou}) is achieved for cars, sofas, and aeroplanes, and the lowest for benches, chairs, and lamps.
Providing the ground-truth poses as supervision improves reconstruction performance in all cases (\itiougt{}).
Note that performance for the concave classes sofa, bathtub, pot, and jar is comparable or higher than several non-concave classes, indicating that our model can indeed  learn them by exploiting shading cues.

Note that in almost all cases, the reconstructed image is very close to the input (\fig{recon-all-classes}); thus, the model has learnt to reconstruct pixels successfully.
Moreover, even when the input is particularly ambiguous due to self-occlusion (e.g. the rightmost car and sofa examples), we see that the model infers a plausible completion of the hidden part of the shape (visible in the third column).
%
However, the subdivision parameterisation limits the amount of detail that can be recovered in some cases, for example the slatted back of the second bench is reconstructed as a continuous surface.
Furthermore, flat surfaces are often reconstructed as several faces that are not exactly coplanar, creating small visual artifacts.
Finally, the use of a fixed-resolution planar mesh limits the smoothness of curved surfaces, as seen in the jar class.

The low values of the pose estimation error \textit{err} (and corresponding high values of \textit{acc}) for most classes indicate that the model has indeed learnt to disentangle pose from shape, without supervision.
This is noteworthy given the model has seen only unannotated 2D images with arbitrary poses ---disentanglement of these factors presumably arises because it is easier for the model to learn to reconstruct in a canonical reference frame, given that it is encouraged by our loss to predict diverse poses.
While the pose estimation appears inaccurate for table, lamp, pot, and jar
note that these classes exhibit rotational symmetry about the vertical axis.
Hence, it is not possible to define (nor indeed to learn) a single, unambiguous canonical frame of reference for them. 
%

\begin{table*}[t]
  \centering
  \caption{
  Reconstruction performance for four classes, with three different mesh parameterisations (\sect{generative}).
  For each class, the first three columns are in the default setting of no pose supervision and correspond to the metrics in \sect{experiments}; \iougt is the IOU when trained with pose supervision. Higher is better for \textit{iou} and \textit{acc}; lower is better for \textit{err}. Experimental setting: single-view training, fixed colour lighting, shading loss.
  }
  \label{tab:class-vs-paramn}

  \begin{tabular}{@{}l @{}>{~~~~~~}c@{} llll @{}>{~~~~~~}c@{} llll @{}>{~~~~~~}c@{} llll @{}>{~~~~~~}c@{} llll@{}}
    \toprule
    && \multicolumn{4}{c}{\textbf{car}} && \multicolumn{4}{c}{\textbf{chair}} && \multicolumn{4}{c}{\textbf{aeroplane}} && \multicolumn{4}{c}{\textbf{sofa}} \\
    \cmidrule{3-6} \cmidrule{8-11} \cmidrule{13-16} \cmidrule{18-21}
    && iou & err & acc & \iougt &&
    iou & err & acc & \iougt &&
    iou & err & acc & \iougt &&
    iou & err & acc & \iougt \\
    ortho-block && 0.72 & 7.6 & \textbf{0.90} & 0.78 && 0.41 & 9.2 & \textbf{0.69} & 0.49 && 0.30 & 7.9 & 0.73 & 0.24 && \textbf{0.59} & 7.3 & \textbf{0.94} & \textbf{0.74} \\
    full-block && 0.54 & 6.5 & 0.82 & 0.63 && \textbf{0.46} & \textbf{4.6} & \textbf{0.69} & \textbf{0.51} && \textbf{0.55} & 1.7 & \textbf{0.90} & 0.57 && 0.39 & 9.1 & 0.70 & 0.68 \\
    subdivision && \textbf{0.77} & \textbf{4.7} & 0.84 & \textbf{0.82} && 0.39 & 7.9 & 0.65 & \textbf{0.51} && \textbf{0.55} & \textbf{1.4} & \textbf{0.90} & \textbf{0.59} && \textbf{0.59} & \textbf{6.5} & 0.88 & 0.71 \\
    \bottomrule
  \end{tabular}
\end{table*}

\begin{table*}[t]
  \centering
  \caption{
  Reconstruction performance with different lighting and loss. \textit{colour} indicates three coloured directional lights with shading loss; \textit{white} indicates a single white directional light plus white ambient, with shading loss; \textit{col+sil} indicates coloured lighting with only the silhouette used in the loss. Our model can exploit the extra information gained by considering shading in the loss, and coloured directional lighting helps further. Experimental setting: single-view training, best mesh parameterisations from \tab{class-vs-paramn}, fixed lighting rotation.
  }
  \label{tab:fixed-lighting}
  \begin{tabular}{@{}l @{}>{~~~~~~}c@{} llll @{}>{~~~~~~}c@{} llll @{}>{~~~~~~}c@{} llll @{}>{~~~~~~}c@{} llll@{}}
    \toprule
    && \multicolumn{4}{c}{\textbf{car}} && \multicolumn{4}{c}{\textbf{chair}} && \multicolumn{4}{c}{\textbf{aeroplane}} && \multicolumn{4}{c}{\textbf{sofa}} \\
    \cmidrule{3-6} \cmidrule{8-11} \cmidrule{13-16} \cmidrule{18-21}
    && iou & err & acc & \iougt &&
    iou & err & acc & \iougt &&
    iou & err & acc & \iougt &&
    iou & err & acc & \iougt \\
    colour && \textbf{0.77} & \textbf{4.7} & \textbf{0.84} & \textbf{0.82} && \textbf{0.46} & \textbf{4.6} & \textbf{0.69} & \textbf{0.51} && \textbf{0.55} & \textbf{1.4} & \textbf{0.90} & \textbf{0.59} && \textbf{0.59} & \textbf{7.3} & \textbf{0.94} & \textbf{0.74} \\
    white && 0.58 & 13.8 & 0.82 & 0.81 && 0.31 & 37.7 & 0.43 & 0.42 && 0.42 & 7.7 & 0.85 & 0.54 && 0.51 & 56.1 & 0.49 & 0.71 \\
    col+sil && 0.46 & 65.2 & 0.29 & 0.64 && 0.28 & 51.7 & 0.35 & 0.48 && 0.20 & 17.8 & 0.57 & 0.47 && 0.27 & 89.8 & 0.15 & 0.57 \\
    \bottomrule
  \end{tabular}
\end{table*}

\subsection{Comparing mesh parameterisations}
\label{sec:experiments/paramns}

We now compare the three mesh parameterisations of \sect{generative}, considering the four classes \textit{car}, \textit{chair}, \textit{aeroplane}, and \textit{sofa}.
We show qualitative results for generation (\fig{gen-different-paramns}) and reconstruction (\fig{recon-different-paramns}); \tab{class-vs-paramn} gives quantitative results for reconstruction.
Again we use fixed colour lighting, shading loss and single-view training.

\begin{sloppypar}
We see that different parameterisations are better suited to different classes, in line with our expectations.
Cars have smoothly curved edges, and are well-approximated by a single simply-connected surface; hence, \textbf{subdivision} performs well.
Conversely, \textbf{ortho-block} fails to represent the curved and non-axis-aligned surfaces, in spite of giving relatively high IOU.
Chairs vary in topology (e.g. the back may be solid or slatted) and sometimes have non-axis-aligned surfaces, so the flexible \textbf{full-block} parameterisation performs best.
Interestingly, \textbf{subdivision} is able to partially reconstruct the holes in the chair backs by deforming the reconstructed surface such that it self-intersects.
Aeroplanes have one dominant topology and include non-axis-aligned surfaces; both \textbf{full-block} and \textbf{subdivision} perform well here.
However, the former sometimes has small gaps between blocks, failing to reflect the true topology.
Sofas often consist of axis-aligned blocks, so the \textbf{ortho-block} parameterisation is expressive enough to model them.
We hypothesise that it performs better than the more flexible \textbf{full-block} as it is easier for training to find a good solution in a more restricted representation space.
This is effectively a form of regularisation.
Overall, the best reconstruction performance is achieved for cars, which accords with \citet{tulsiani17cvpr}, \citet{yan16nips}, and \citet{fan17cvpr}.
%
On average over the four classes, the best parameterisation is \textbf{subdivision}, both with and without pose supervision.
\end{sloppypar}

\begin{table*}[t]
  \centering
  \caption{
  Reconstruction performance with fixed and varying lighting. In the \textit{varying} case, our model must learn to predict the lighting angle, simultaneously with exploiting the shading cues it provides. Experimental setting: single-view training, best mesh parameterisations from \tab{class-vs-paramn}, shading loss.
  }
  \label{tab:varying-lighting}
  \begin{tabular}{@{}l@{~~}l @{}>{~~~~~}c@{} llll @{}>{~~~~~}c@{} llll @{}>{~~~~~}c@{} llll @{}>{~~~~~}c@{} llll@{}}
    \toprule
    &&& \multicolumn{4}{c}{\textbf{car}} && \multicolumn{4}{c}{\textbf{chair}} && \multicolumn{4}{c}{\textbf{aeroplane}} && \multicolumn{4}{c}{\textbf{sofa}} \\
    \cmidrule{4-7} \cmidrule{9-12} \cmidrule{14-17} \cmidrule{19-22}
    &&& iou & err & acc & \iougt &&
    iou & err & acc & \iougt &&
    iou & err & acc & \iougt &&
    iou & err & acc & \iougt \\
    fixed & white && 0.58 & 13.8 & 0.82 & 0.81 && 0.31 & 37.7 & 0.43 & 0.42 && 0.42 & 7.7 & 0.85 & 0.54 && 0.51 & 56.1 & 0.49 & 0.71 \\
    varying & white && 0.48 & 23.6 & 0.58 & 0.79 && 0.31 & 31.1 & 0.47 & 0.43 && 0.40 & 2.5 & 0.82 & 0.55 && 0.47 & 60.7 & 0.47 & 0.71 \\
    fixed & colour && 0.77 & 4.7 & 0.84 & 0.82 && 0.46 & 4.6 & 0.69 & 0.51 && 0.55 & 1.4 & 0.90 & 0.59 && 0.59 & 7.3 & 0.94 & 0.74 \\
    varying & colour && 0.60 & 10.5 & 0.82 & 0.79 && 0.32 & 36.5 & 0.42 & 0.46 && 0.52 & 2.4 & 0.89 & 0.59 && 0.69 & 7.5 & 0.96 & 0.73 \\
    \bottomrule
  \end{tabular}
\end{table*}

\subsection{Lighting}
\label{sec:experiments/lighting}

\paragraph{Fixed lighting rotation.}
\tab{fixed-lighting} shows how reconstruction performance varies with the different choices of lighting, \textbf{colour} and \textbf{white}, using \textbf{shading} loss.
Coloured directional lighting provides more information during training than white lighting, and the results are correspondingly better.

\begin{sloppypar}
We also show performance with \textbf{silhouette} loss for coloured light.
This considers just the silhouette in the reconstruction loss, instead of the shaded pixels.
To implement it, we differentiably binarise both our reconstructed pixels $I_0$ and the ground-truth pixels $\mathbf{x}^{(i)}$ prior to calculating the reconstruction loss.
Specifically, we transform each pixel $p$ into $p / (p + \eta)$, where $\eta$ is a small constant.
This performs significantly worse than with shading in the loss, in spite of the input images being identical.
Thus, back-propagating information from shading through the renderer does indeed help with learning---it is not merely that colour images contain more information for the encoder network.
%
As in the previous experiment, we see that pose supervision helps the model (column \itiougt{} vs. \textit{iou}).
In particular, only with pose supervision are silhouettes informative enough for the model to learn a canonical frame of reference reliably, as evidenced by the high median rotation errors without (column \textit{err}).
\end{sloppypar}

\paragraph{Varying lighting rotation.}
We have shown that shading cues are helpful for training our model. We now evaluate whether it can still learn successfully when the lighting angle varies across training samples (\textbf{varying}).
\tab{varying-lighting} shows that our method can indeed reconstruct shapes even in this case.
When the object pose is given as supervision (column \itiougt{}), the reconstruction accuracy is on average only slightly lower than in the case of fixed, known lighting.
Thus, the encoder successfully learns to disentangle the lighting angle from the surface normal orientation, while still exploiting the shading information to aid reconstruction.
When the object pose is not given as supervision (column \textit{iou}), the model must learn to simultaneously disentangle shape, pose and lighting.
Interestingly, even in this extremely hard setting our method still manages to produce good reconstructions, although of course the accuracy is usually lower than with fixed lighting.
%
Finally, note that our results with varying lighting are better than those with fixed lighting from the final row of \tab{fixed-lighting}, using only the silhouette in the reconstruction loss.
This demonstrates that even when the model does not have access to the lighting parameters, it still learns to benefit from shading cues, rather than simply using the silhouette.

\begin{figure}
  \centering
  \includegraphics[width=\linewidth,trim={2cm 2cm 2cm 2cm},clip]{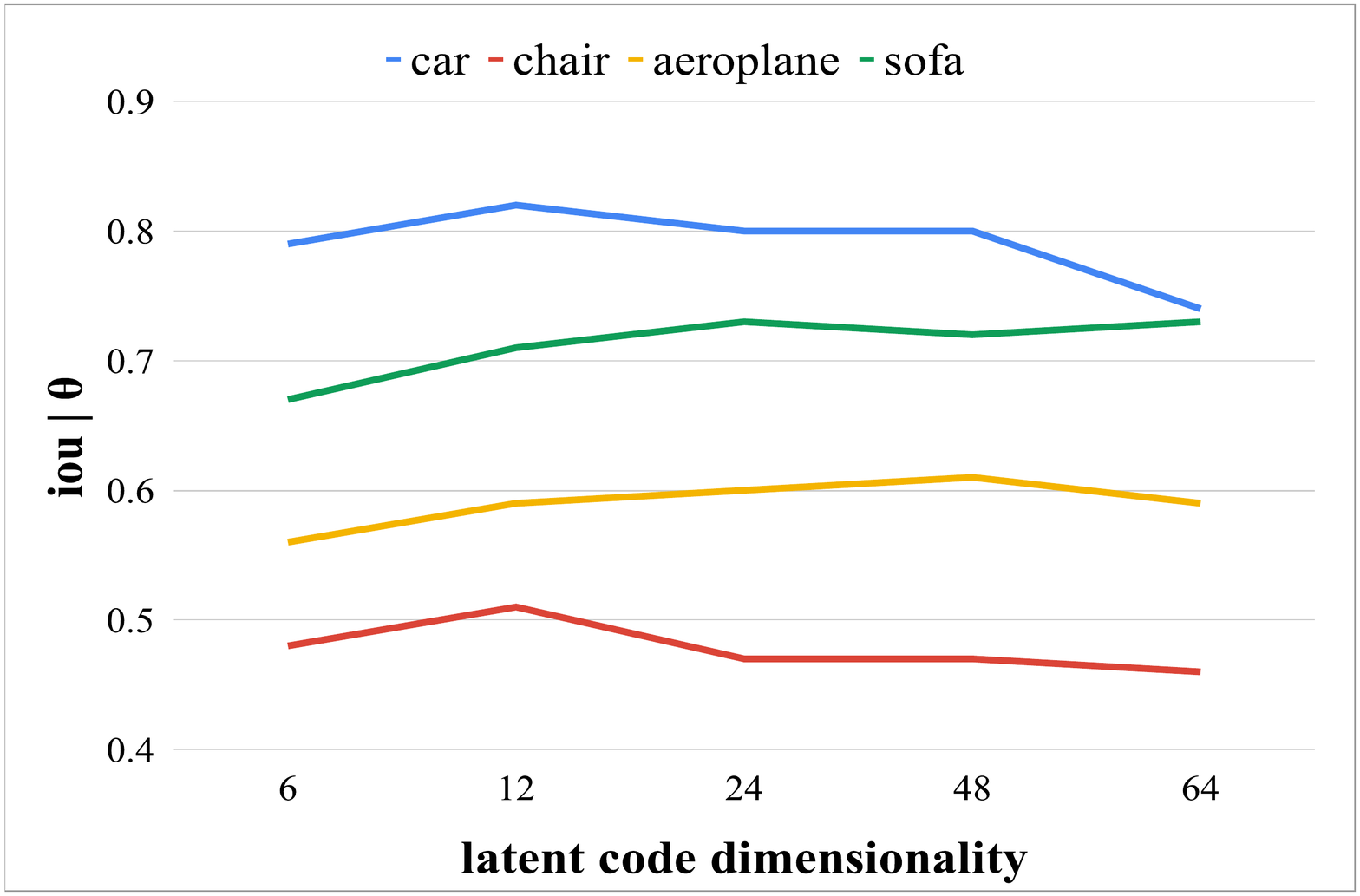}
  \caption{  Effect of varying the dimensionality of the latent embedding vector $\mathbf{z}$ on reconstruction performance (\iougt). Experimental setting: subdivision, fixed colour lighting, shading loss.
  }
  \label{fig:latent-dimensionality}
\end{figure}

\begin{figure}
  \centering
  \includegraphics[width=\linewidth]{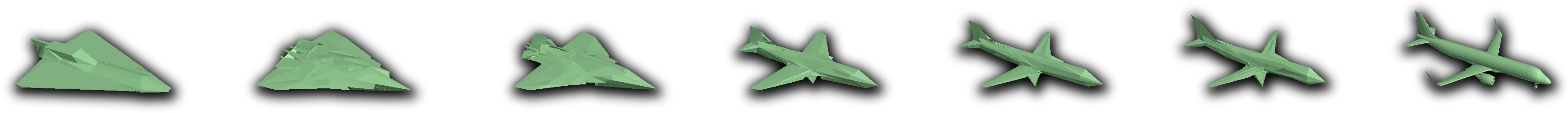}\\
  \includegraphics[width=\linewidth]{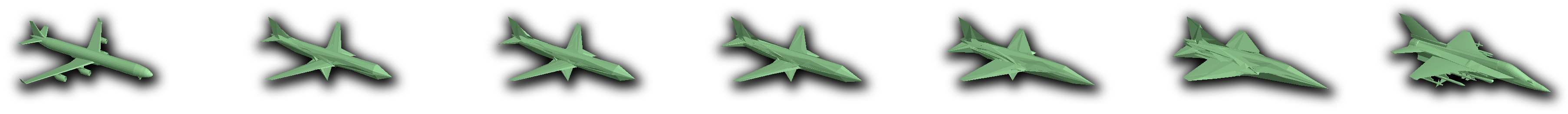}\\
  \includegraphics[width=\linewidth]{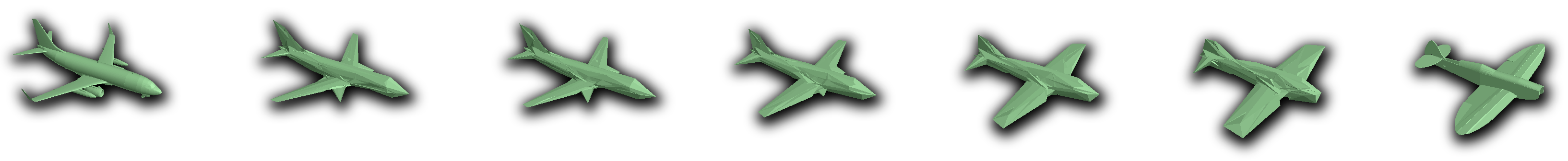}\\
  \includegraphics[width=\linewidth]{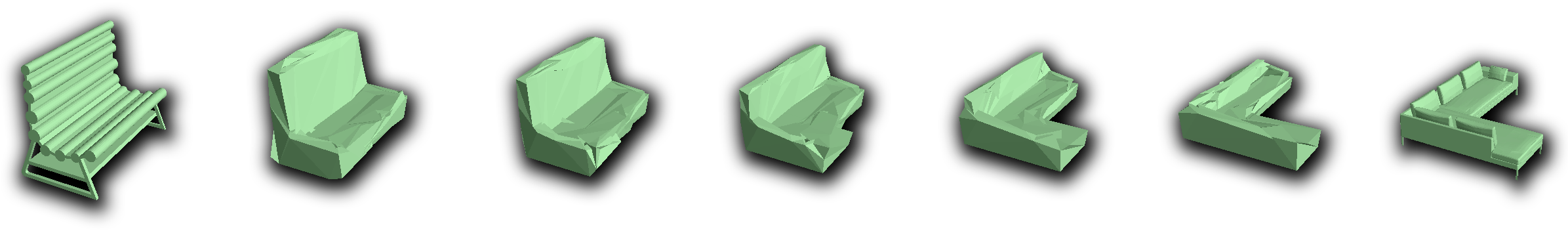}\\
  \includegraphics[width=\linewidth]{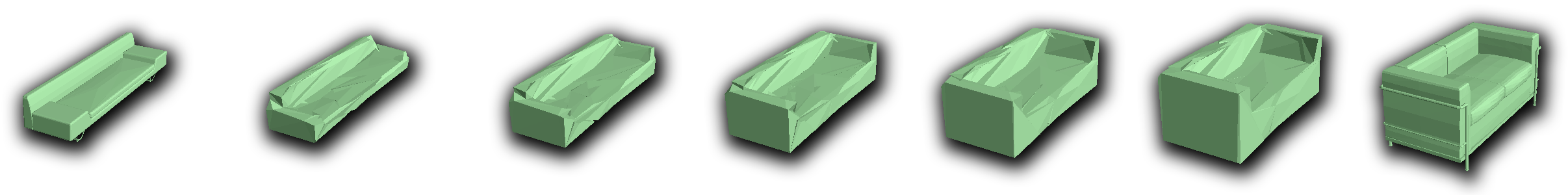}\\
  \includegraphics[width=\linewidth]{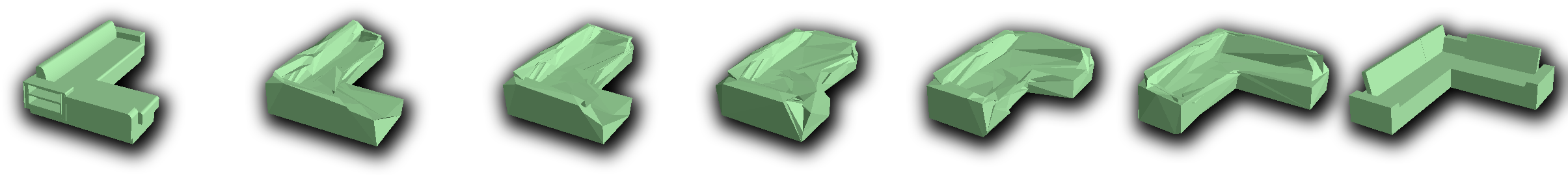}
  \caption{
  Interpolating between shapes in latent space.
  In each row, the leftmost and rightmost images show ground-truth shapes from ShapeNet, and the adjacent columns show the result of reconstructing each using our model with \textbf{subdivision} parameterisation.
  In the centre three columns, we interpolate between the resulting latent embeddings, and display the decoded shapes.
  In each case, we see a semantically-plausible, gradual deformation of one shape into the other.
  }
  \label{fig:latent-interpolations}
\end{figure}

\begin{table}
  \centering
  \caption{
  Reconstruction performance with multiple views at train/test time. Our model is able to exploit the extra information gained through multiple views, and can benefit even when testing with a single view. Experimental setting: best mesh parameterisations from \tab{class-vs-paramn}, fixed colour lighting, shading loss.
  }
  \label{tab:multi-view}
  \begin{tabular}{@{}c@{~~}c @{}>{~~~~}c@{} llll @{}>{~~~~}c@{} llll@{}}
    \toprule
    \multicolumn{2}{c}{\textbf{views}} && \multicolumn{4}{c}{\textbf{car}} && \multicolumn{4}{c}{\textbf{chair}} \\
    \cmidrule{4-7} \cmidrule{9-12}
    train & test && iou & err & acc & \iougt &&
    iou & err & acc & \iougt \\
    1 & 1 && 0.77 & 4.7 & 0.84 & 0.82 && 0.46 & 4.6 & 0.69 & 0.51 \\
    3 & 1 && 0.82 & \textbf{1.3} & \textbf{0.94} & 0.83 && 0.50 & \textbf{2.1} & \textbf{0.83} & 0.52 \\
    3 & 3 && \textbf{0.83} & 1.7 & \textbf{0.94} & \textbf{0.84} && \textbf{0.53} & 3.1 & 0.80 & \textbf{0.56} \\
    \bottomrule
  \end{tabular}
\end{table}

\begin{table*}
  \centering
  \caption{
  Reconstruction performance (\iougt) in a setting matching \citet{yan16nips}, \citet{tulsiani17cvpr}, \citet{kato18cvpr}, and \citet{yang18eccv}, which are silhouette-based methods trained with pose supervision and multiple views (to be precise, \citet{yang18eccv} provide pose annotations for 50\% of all training images).
	\textit{PTN, our images} is running the unmodified public code of \citet{yan16nips} with their normal silhouette loss, on our coloured images.
  $N_\text{views}$ indicates the number of views of each instance provided together in each minibatch during training.
  The final rows show performance of two state-of-the-art methods with full 3D supervision~\citep{fan17cvpr,richter18cvpr}---note that our colour results are comparable with these, in spite of using only 2D images.
  Experimental setting: subdivision, three views per object during training, fixed lighting rotation.
  }
  \label{tab:fixed-pose-competitors}
  \begin{tabular}{@{}l ccc cccc@{}}
    \toprule
    & $N_\text{views}$ & lighting & loss & car & chair & aeroplane & sofa \\
    \midrule
    PTN~\citep{yan16nips} & 24 & white & silhouette & 0.71 & \textbf{0.50} & 0.56 & 0.62 \\
    DRC~\citep{tulsiani17cvpr} & 5 & white & silhouette & 0.73 & 0.43 & 0.50 & - \\
    DRC~\citep{tulsiani17cvpr} & 5 & white & depth & 0.74 & 0.44 & 0.49 & - \\
    NMR~\citep{kato18cvpr} & 2 & white & silhouette & 0.71 & \textbf{0.50} & \textbf{0.62} & 0.67 \\
    LPS~\citep{yang18eccv} & 2 & white & silhouette & 0.78 & 0.44 & 0.57 & 0.54 \\
    \midrule
    PTN, our images & 24 & colour & silhouette & 0.66 & 0.22 & 0.42 & 0.46 \\
    \midrule
    ours & 3 & white & silhouette & 0.79 & 0.46 & 0.58 & 0.67 \\
    ours & 3 & white & shading & 0.81 & 0.48 & 0.60 & 0.67 \\
    ours & 3 & colour & shading & \textbf{0.83} & \textbf{0.50} & 0.61 & \textbf{0.73} \\
    \midrule
    \textit{PSG~\citep{fan17cvpr}} & - & \textit{white} & \textit{3D} & \textit{0.83} & \textit{0.54} & \textit{0.60} & \textit{0.71} \\
    \textit{MN~\citep{richter18cvpr}} & - & \textit{white} & \textit{3D} & \textit{0.85} & \textit{0.55} & \textit{0.65} & \textit{0.68} \\
    \bottomrule
  \end{tabular}
\end{table*}

\subsection{Latent space structure}
\label{sec:experiments/latent-space}

The shape of a specific object instance must be entirely captured by the latent embedding vector $\mathbf{z}$.
On the one hand, using a higher dimensionality for $\mathbf{z}$ should result in better reconstructions, due to the greater representational power.
On the other hand, a lower dimensionality makes it easier for the model to learn to map \textit{any} point in $\mathbf{z}$ to a reasonable shape, and to avoid over-fitting the training set.
To evaluate this trade-off, we ran experiments with different dimensionalities for $\mathbf{z}$ (\fig{latent-dimensionality}).
We see that for all classes, increasing from 6 to 12 dimensions improves reconstruction performance on the test set.
Beyond 12 dimensions, the effect differs between classes.
For car and chair, higher dimensionalities yield lower performance (indicating over-fitting or other training difficulties). Instead, aeroplane and sofa continue to benefit from higher and higher dimensionalities, up to 48 for aeroplane and 64 (and maybe beyond) for sofa.

For all our other experiments, we use a 12-dimensional embedding, as this gives good performance on average across classes.
Note that our embedding dimensionality is much smaller than its counterpart in other works.
For example, \citet{tulsiani17cvpr} have a bottleneck layer with dimensionality 100, while \citet{wiles17bmvc} use dimensionality 160.
This low dimensionality of our embeddings facilitates the encoder mapping images to a compact region of the embedding space centred at the origin; this in turn allows modelling the embeddings by a simple Gaussian from which samples can be drawn.

\paragraph{Interpolating in the latent space.}
To demonstrate that our models have learnt a well-behaved manifold of shapes for each class, we select pairs of ground-truth shapes, reconstruct these using our model, and linearly interpolate between their latent embeddings (\fig{latent-interpolations}).
We see that the resulting intermediate shapes give a gradual, smooth deformation of one shape into the other, showing that all regions of latent space that we traverse correspond to realistic samples.

\subsection{Multi-view training/testing}
\label{sec:experiments/multi-view}

\tab{multi-view} shows results when we provide multiple views of each object instance to the model, either at training time only, or during both training and testing.
In both cases, this improves results over using just a single view---the model has learnt to exploit the additional information about each instance.
Note that when training with three views but testing with one, the network has not been optimised for the single-view task; however, the additional information present during training means it has learnt a stronger model of valid shapes, and this knowledge transfers to the test-time scenario of reconstruction from a single image.

\subsection{Comparison to previous and concurrent works}
\label{sec:experiments/prior-comparison}

\begin{figure}
  \centering
  \includegraphics[width=0.9\linewidth]{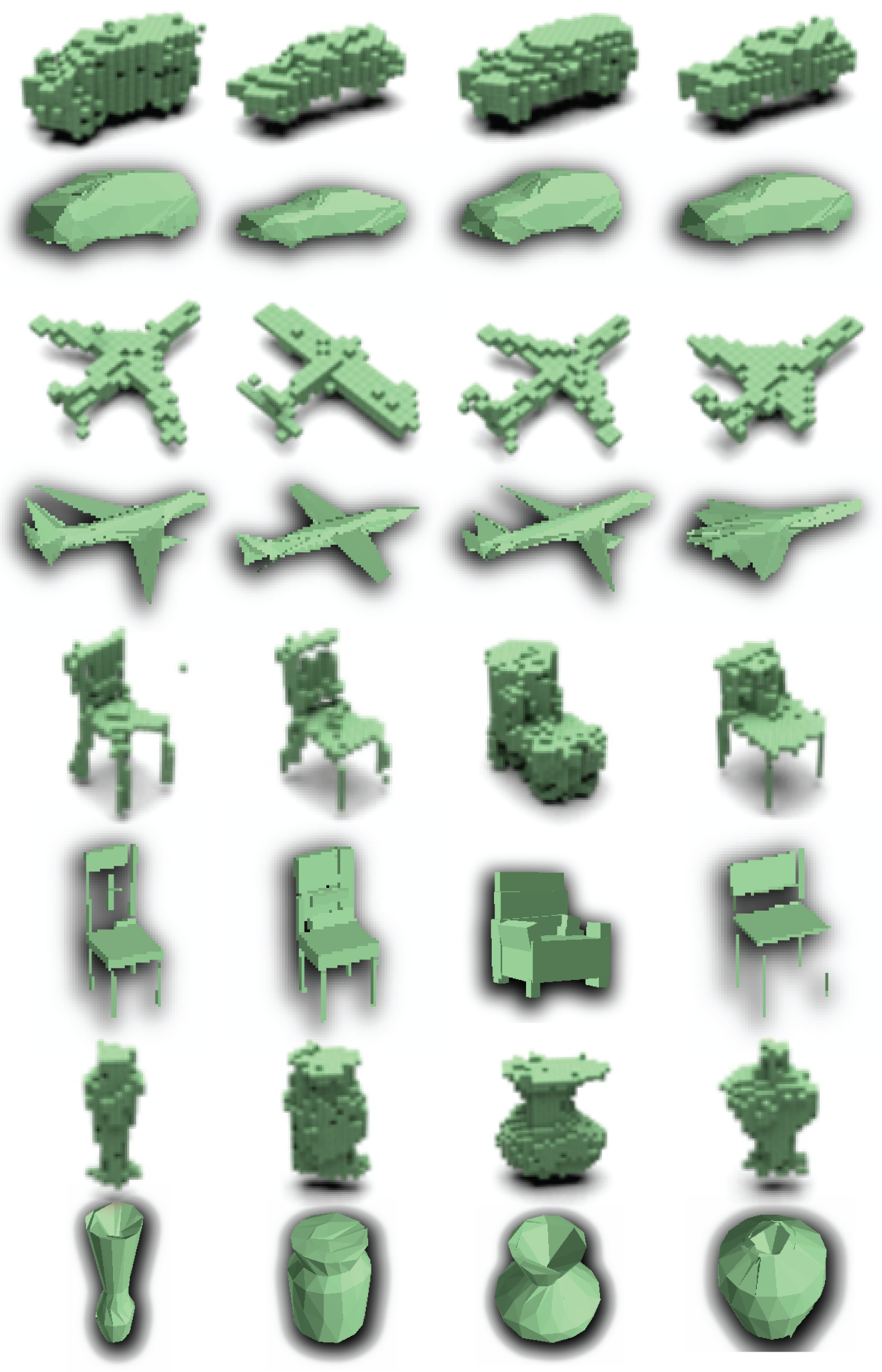}
  \caption{
  Samples from the voxel-based method of \citet{gadelha173dv} (odd rows), shown above stylistically-similar samples from our model (even rows).
  Both methods are trained with a single view per instance, and without pose annotations. However, our model outputs meshes, and uses shading in the loss; hence, it can represent smooth surfaces and learn concave classes such as \textit{vase}.
  }
  \label{fig:samples-vs-gadelha}
\end{figure}

\paragraph{Generation.}
\fig{samples-vs-gadelha} compares samples from our model, to samples from that of \citet{gadelha173dv}, on the four object classes we have in common.
This is the only prior work that trains a 3D generative model using only single views of instances, and without pose supervision.
Note however that unlike us, all images in the training set of \citet{gadelha173dv} are taken from one of a fixed set of eight poses, making their task a little easier.
We manually selected samples from our model that are stylistically similar to those shown in \citet{gadelha173dv} to allow side-by-side comparison.
We see that in all cases, generating meshes tends to give cleaner, more visually-pleasing samples than their use of voxels.
For \textit{chair}, our model is able to capture the very narrow legs; for \textit{aeroplane}, it captures the diagonal edges of the wings; for \textit{car} and \textit{vase}, it captures the smoothly curved edges.
Note that as shown in \fig{gen-all-classes}, our model also successfully learns models for concave classes such as \textit{bathtub} and \textit{sofa}---which is impossible for \citet{gadelha173dv} as they do not consider shading.

\begin{table*}
  \centering
  \caption{
    Comparison of our method with the concurrent work MVC~\citep{tulsiani18cvpr-mvc} in different settings, on the three classes for which they report results.
    Note that they vary elevation as well as azimuth, and their images are rendered with texturing under white light; hence, this comparison to our method is only approximate.
    Experimental setting: subdivision, three views per object during training, fixed lighting rotation.
  }
  \label{tab:mvc-comparison}
  \begin{tabular}{@{}l ll @{}>{~~~~~~}c@{} llll @{}>{~~~~~~}c@{} llll @{}>{~~~~~~}c@{} llll@{}}
    \toprule
    & \multirow{2}{*}{\textbf{lighting}} & \multirow{2}{*}{\textbf{loss}} && \multicolumn{4}{c}{\textbf{car}} && \multicolumn{4}{c}{\textbf{chair}} && \multicolumn{4}{c}{\textbf{aeroplane}} \\
    \cmidrule{5-8} \cmidrule{10-13} \cmidrule{15-18}
    & & && iou & err & acc & \iougt &&
    iou & err & acc & \iougt &&
    iou & err & acc & \iougt \\
    ours & white & silhouette && 0.62 & 19.4 & 0.55 & 0.79 && 0.45 & 13.1 & 0.60 & 0.46 && 0.56 & 1.4 & 0.83 & 0.58 \\
    ours & white & shading && 0.77 & 3.0 & 0.91 & 0.81 && 0.46 & 4.2 & \textbf{0.83} & 0.48 && 0.57 & 1.0 & \textbf{0.89} & 0.60 \\
    ours & colour & shading && \textbf{0.82} & \textbf{1.3} & \textbf{0.94} & \textbf{0.83} && \textbf{0.47} & \textbf{2.7} & 0.82 & \textbf{0.50} && \textbf{0.58} & \textbf{0.9} & 0.88 & \textbf{0.61} \\
    MVC & white & silhouette && 0.74 & 5.2 & 0.87 & 0.75 && 0.40 & 7.8 & 0.81 & 0.42 && 0.52 & 14.3 & 0.69 & 0.55 \\
    MVC & white & depth && 0.71 & 4.9 & 0.85 & 0.69 && 0.43 & 8.6 & \textbf{0.83} & 0.45 && 0.44 & 21.7 & 0.60 & 0.43 \\
    \bottomrule
  \end{tabular}
\end{table*}

\paragraph{Reconstruction.}
\tab{fixed-pose-competitors} compares our results with previous and concurrent 2D-supervised methods that input object pose at training time.
We consider works that appeared in 2018 to be concurrent to ours~\citep{henderson18bmvc}.
Here, we conduct experiments in a setting matching \citet{yan16nips}, \citet{tulsiani17cvpr}, \citet{kato18cvpr}, and \citet{yang18eccv}: multiple views at training time, with ground-truth pose supervision (given for 50\% of images in~\citet{yang18eccv}).

Even when using only silhouettes during training, our results are about as good as the best of the works we compare to, that of \citet{kato18cvpr}, which is a concurrent work.
Our results are somewhat worse than theirs for aeroplanes and chairs, better for cars, and identical for sofas. On average over the four classes, we reach the same iou of 62.5\%.
When we add shading information to the loss, our results show a significant improvement. Importantly, \citet{yan16nips}, \citet{tulsiani17cvpr} and \citet{yang18eccv} cannot exploit shading, as they are based on voxels.
%
Coloured lighting helps all classes even further, leading to a final performance higher than than all other methods on car and sofa, and comparable to the best other method on chair and aeroplane~\citep{kato18cvpr}. On average we reach 66.8\% iou, compared to 62.5\% for \citet{kato18cvpr}.

\begin{sloppypar}
We also show results for \citet{yan16nips} using our coloured lighting images as input, but their silhouette loss\footnote{we use their publicly-available implementation from \url{https://github.com/xcyan/nips16_PTN}, unmodified apart from changing the camera parameters to match our renderings}.
This performs worse than our method on the same images, again showing that incorporating shading in the loss is useful---our colour images are not simply more informative to the encoder network than those of \citet{yan16nips}.
Interestingly, when trained with shading or colour, our method outperforms \citet{tulsiani17cvpr} even when the latter is trained with depth information.
When trained with colour, our results (average 66.8\% iou) are even close to those of \citet{fan17cvpr} (67.0\%) and \citet{richter18cvpr} (68.2\%), which are state-of-the-art methods trained with full 3D supervision.
\end{sloppypar}

\tab{mvc-comparison} compares our results with those of \citet{tulsiani18cvpr-mvc}.
This is a concurrent work similar in spirit to our own, that learns reconstruction and pose estimation without 3D supervision nor pose annotations, but requires multiple views of each instance to be presented together during training.
We match their experimental setting by training our models on three views per instance; however, they vary elevation as well as azimuth during training, making their task a little harder.
%
We see that the ability of our model to exploit shading cues enables it to significantly outperform \citet{tulsiani18cvpr-mvc}, which relies on silhouettes in its loss.
This is shown by \textit{iou} and \itiougt{} being higher for our method with white light and shading loss, than for theirs with white light and silhouette.
Indeed, our method outperforms theirs even when they use depth information as supervision.
When we use colour lighting, our performance is even higher, due to the stronger information about surface normals.
Conversely, when our method is restricted to silhouettes, it performs significantly worse than theirs across all three object classes.

\section{Conclusion}
We have presented a framework for generation and reconstruction of 3D meshes.
Our approach is flexible and supports many different supervision settings, including weaker supervision than any prior works (i.e. a single view per training instance, and without pose annotations).
When pose supervision is not provided, it automatically learns to disentangle the effects of shape and pose on the final image.
When the lighting is unknown, it also learns to disentangle the effects of lighting and surface orientation on the shaded pixels.
%
We have shown that exploiting shading cues leads to higher performance than state-of-the-art methods based on silhouettes~\citep{kato18cvpr}.
It also allows our model to learn concave classes, unlike these prior works.
Moreover, our performance is higher than that of methods with depth supervision~\citep{tulsiani17cvpr,tulsiani18cvpr-mvc}, and even close to the state-of-the-art results using full 3D supervision~\citep{fan17cvpr,richter18cvpr}.
Finally, ours is the first method that can learn a generative model of 3D meshes, trained with only 2D images.
We have shown that use of meshes leads to more visually-pleasing results than prior voxel-based works \citep{gadelha173dv}.

\paragraph{Limitations.}
%
%
Our method is trained to ensure that the rendered reconstructions match the original images.
Such an approach is inherently limited by the requirement that images from the generative model must resemble the input images for reconstruction, in terms of the L2 distance on pixels.
Thus, in order to operate successfully on natural images, the model would need to be extended to incorporate more realistic materials and lighting. 

Our use of different mesh parameterisations gives flexibility to model different classes faithfully.
We have shown that the subdivision parameterisation gives reasonable results for all classes; however, other parameterisations work better for particular classes.
Hence, for best results on a given class, a suitable parameterisation must be selected by the user.

Finally, we note that when multiple views but only silhouettes are available as input, discriminative methods specialised for this task \citep{kato18cvpr,tulsiani18cvpr-mvc} outperform our approach.

\appendix

\section{Network architectures}
\label{app:net-arch}

In this appendix we briefly describe the architectures of the decoder and encoder neural networks.

The decoder network $F_\phi$ takes the latent embedding $\mathbf{z}$ as input.
This is passed through a fully-connected layer with 32 output channels using ReLU activation.
The resulting embedding is processed by a second fully-connected layer that outputs the mesh parameters: vertex offsets for subdivision parameterisation, and locations, scales and rotations for the primitive-based parameterisations.
For the primitive scales, we use a softplus activation to ensure they are positive; for the other parameters, we do not use any activation function.

The encoder network $\mathrm{enc}_\omega(\mathbf{x})$ is a CNN operating on RGB images of size $128 \times 96$ pixels; its architecture is similar to that of \citet{wiles17bmvc}.
Specifically, it has the following layers, each with batch normalisation and ReLU activation:
\begin{itemize}[noitemsep,nolistsep]
  \item  $3 \times 3$ convolution, 32 channels, stride = 2
  \item  $3 \times 3$ convolution, 64 channels, stride = 1
  \item $2 \times 2$ max-pooling, stride = 2
  \item  $3 \times 3$ convolution, 96 channels, stride = 1
  \item $2 \times 2$ max-pooling, stride = 2
  \item $3 \times 3$ convolution, 128 channels, stride = 1
  \item $2 \times 2$ max-pooling, stride = 2
  \item $4 \times 4$ convolution, 128 channels, stride = 1
  \item fully-connected, 128 channels
\end{itemize}
This yields a 128-dimensional feature vector for the image.
The parameters for each variational distribution are produced by a further fully-connected layer, each taking this feature vector as input.
For the mean of $\mathbf{z}$, we do not use any activation function; for the mean of $\theta_\mathrm{fine}$ we use tanh activation, scaled by $\pi/R_\theta$ to ensure $\theta_\mathrm{coarse}$ rather than $\theta_\mathrm{fine}$ is used to model large rotations.
For the mean of $\lambda_\mathrm{fine}$ we analogously use tanh activation scaled by $\pi/R_\lambda$.
For the standard deviations of $\mathbf{z}$, $\theta_\mathrm{fine}$, and $\lambda_\mathrm{fine}$, we use softplus activation, to ensure they are positive.
Finally, for $\theta_\mathrm{coarse}$ and $\lambda_\mathrm{coarse}$, we use softmax outputs giving the probabilities of the different coarse rotations.

\section{Hyperparameters}
\label{app:hyperparams}

We now give the values for the hyperparameters defined in \sect{generative} and \sect{training}, that we used when training our models.
\begin{itemize}
  \item learning rate: $10^{-3}$ (constant throughout training), apart from parameters specific to rotation decoder in full-block parameterisation, for which we used $10^{-4}$
  \item gradient clipping: global Euclidean norm at most 5
  \item KL loss weight $\beta = 10^3$
  \item discrete prior-matching loss weight $\alpha = 5 \times 10^5$
  \item number of object azimuth bins $R_\theta = 12$
  \item number of lighting azimuth bins $R_\lambda = 3$, covering only the interval $[0,\,\pi)$ as the renderer uses double-sided lighting calculations
  \item numbers of blocks in primitive-based parameterisations: 6 for ortho-block, 12 for full-block
  \item number of subdivisions in subdivision parameterisation: 4 segments along each axis
\end{itemize}


 \bibliographystyle{spbasic}      
\bibliography{/home/phenders/bibtex/longstrings,/home/phenders/bibtex/calvin}   

\end{document}